\documentclass[journal]{IEEEtran}
\ifCLASSINFOpdf
\else
\fi
\ifCLASSOPTIONcompsoc
  \usepackage[caption=false,font=normalsize,labelfont=sf,textfont=sf]{subfig}
\else
  \usepackage[caption=false,font=footnotesize]{subfig}
\fi

\usepackage{stfloats}
\usepackage{graphicx,import}
\usepackage{epsfig} 
\usepackage{mathptmx} 
\usepackage{times} 
\usepackage{amsmath} 
\usepackage{amssymb}  
\usepackage{epstopdf}
\usepackage{algorithmicx}
\usepackage{algorithm}
\usepackage[noend]{algpseudocode}
\usepackage{cleveref}

\usepackage{enumitem}
\setenumerate{leftmargin=*}
\usepackage{marvosym}
\usepackage{amssymb}
\usepackage[usenames, dvipsnames]{xcolor, colortbl}
\usepackage{cancel}
\usepackage{xcolor}
\usepackage{balance}
\usepackage{minibox}
\usepackage{units}
\usepackage{bm}

\newcommand{\beq}{\begin{equation}}
\newcommand{\eeq}{\end{equation}}
\newcommand{\beqo}{\begin{displaymath}}
\newcommand{\eeqo}{\end{displaymath}}

\newcommand{\bea}[1]{\begin{align}#1\end{align}}
 
\newcommand{\beao}{\begin{eqnarray*}}
\newcommand{\eeao}{\end{eqnarray*}}

\newcommand{\deltaxvec}[1]{{\scriptsize \Delta}\xvec}

\newcommand{\qvec}{{\mathbf{q}}}

\newcommand{\xvec}{{\mathbf{x}}}

\newcommand\path{\chi}

\newcommand\trajectory{\xvec}

\newcommand\idea[4]{}

\usepackage{tikz}
\usepackage[color=black,opacity=1,angle=0,scale=1]{background}
\backgroundsetup{
contents={
\begin{tikzpicture}
\node at (current page.center) [align=center] {\textcopyright 2018 IEEE. Personal use of this material is permitted. 
\\ Permission from IEEE must be obtained for all other uses, in any current or future media, including reprinting/republishing this material for advertising or promotional purposes, 
\\ creating new collective works, for resale or redistribution to servers or lists, or reuse of any copyrighted component of this work in other works.
\\ DOI: 10.1109/MITS.2018.2867525};
\end{tikzpicture}},
placement=bottom,
scale=0.6,
vshift=20
}

\begin{document}
%
\title{
Intersection Warning System for Occlusion Risks using Relational Local Dynamic Maps
}

%
%
%

\author{Florian Damerow$^{1\text{\fontsize{10}{12}\selectfont *}}$, Tim Puphal$^{2\text{\fontsize{10}{12}\selectfont *}}$, Benedict Flade$^{2}$, Yuda Li$^{1}$ and Julian Eggert$^{2}$
\thanks{$^{1}$ Florian Damerow and Yuda Li were with the Control Methods and Robotics Lab, Technical University of Darmstadt, 64283 Darmstadt, Germany
        {\tt\small florian.damerow@rmr.tu-darmstadt.de}}%
\thanks{$^{2}$ Tim Puphal, Benedict Flade and Julian Eggert are with the Honda Research Institute (HRI) Europe, 
	Carl-Legien-Str. 30, 63073 Offenbach, Germany {\tt\small [tim.puphal, benedict.flade, julian.eggert]@honda-ri.de}}
\thanks{$\text{\fontsize{10}{12}\selectfont *}$ The authors contributed equally to this work}
}

\maketitle

\begin{abstract}
This work addresses the task of risk evaluation in traffic scenarios with limited observability due to restricted sensorial coverage.
Here, we concentrate on intersection scenarios that are difficult to access visually. 
To identify the area of sight, we employ ray casting on a local dynamic map providing geometrical information and road infrastructure. Based on the area with reduced visibility, 
we first model scene entities that pose a potential risk without being visually perceivable yet. Then, we predict a worst-case trajectory in the survival analysis for collision risk estimation. 
Resulting risk indicators are utilized to evaluate the driver's current behavior, to warn the driver in critical situations, to give suggestions on how to act safely or to plan safe trajectories.
We validate our approach by applying the resulting intersection warning system on real world scenarios.  
The proposed system's behavior reveals to mimic the general behavior of a correctly acting human driver.
\end{abstract}


%
\IEEEpeerreviewmaketitle

%
%
%

\section{Introduction} 
\IEEEPARstart{I}{n} recent years a wide range of advanced driver assistance systems (ADAS) has been developed in order to detect upcoming obstacles and avoid collisions. 
In general, static and dynamic objects front of the vehicle are captured by on-board sensors, such as lidar, radar or camera.
Based on scene observations, upcoming hazards are determined to warn the driver or actively perform an evasive maneuver. 
As an example, adaptive cruise control (ACC) systems \cite{winner2015acc} detect leading vehicles and adapt the ego driver's velocity to safely follow the leading vehicle.

State of the art intersection warning systems identify collisions purely based on the actual sensor input, but do not consider sensor limitations due to occlusion. 
In this sense, research presented in \cite{aycard2011intwarn} 
predicts trajectories of all traffic participants to determine their intersecting points and 
calculates the corresponding risk employing the time-to-collision (TTC) indicator. 

To overcome sensor range limitations, recent approaches, such as \cite{andrews1999intwarn}, rely on active systems pysically integrated in the road infrastructure 
at intersections. Approaching vehicles that are undetectable by on-board sensors, are detected by the road infrastructure unit, e.g. by vehicle-to-infrastructure (V2I) 
communication, to warn all drivers of imminent collisions. 
However, such infrastructure units are mostly not present and the majority of nowadays traffic participants is not equipped with such communication systems.

We propose a method to assess the risk with on-board sensors of the ego vehicle. Hereby we focus on scenarios in which surrounding buildings hinder the perception of approaching traffic participants. 
For this purpose, we (1) estimate critical occluded areas, 
(2) model virtual vehicles with specific behaviors and (3) calculate inherent risk
for the ego vehicle. 
This method is then applied to evaluate the driver's current behavior when approaching an intersection with areas of limited visibility in order to issue a warning in critical situations. Our intersection warning system considers sensor limitations 
without the restraint of requiring additional hardware. The prediction of occlusion risk was presented previously in \cite{intwarn2017}. In this paper, it is extended by a relational local dynamic map (R-LDM) \cite{ldm2017} and evaluated in more extensive experiments.

\subsection{Related Work} \label{sec:related}

\noindent Uniform occlusions in traffic scenes, covering the whole or large parts of the image, can arise from weather conditions (dust, fog, rain and snow) 
or sun paths (brightness, darkness or shadows). 
The visibility reduction caused by fog is shown in \cite{pomerleau1998}. The authors use camera data in order to exploit the difference of intensity in lane markings
at various distances ahead of the vehicle. 
Similarly, \cite{kimura2007} derived metrics for the detectability and discriminality of traffic lights in images with not equally distributed illumination..
In both cases, occlusions affect the overall visibility. 

On the contrary, sources for selective occlusions 
are dynamic entities (e.g. oversized vehicles) and static objects (e.g. buildings and trees).
They can be incorporated in the driver's sensory observable area in multiple ways. For instance, \cite{seo2008} uses a lidar sensor to extract a three dimensional model of the environment, which is transformed into an object-free planar area. 
Furthermore with a stereo camera, object detection algorithms \cite{sivaraman2013} are utilized to filter cars which reduce the visible area. 
\cite{pepik2013} focuses on detecting occlusion patterns in cars 
using support vector machines (SVM).

\begin{figure*}[t!]
      \centering
      \parbox{1.0\linewidth}{
      \centering
      \includegraphics[width=\linewidth]{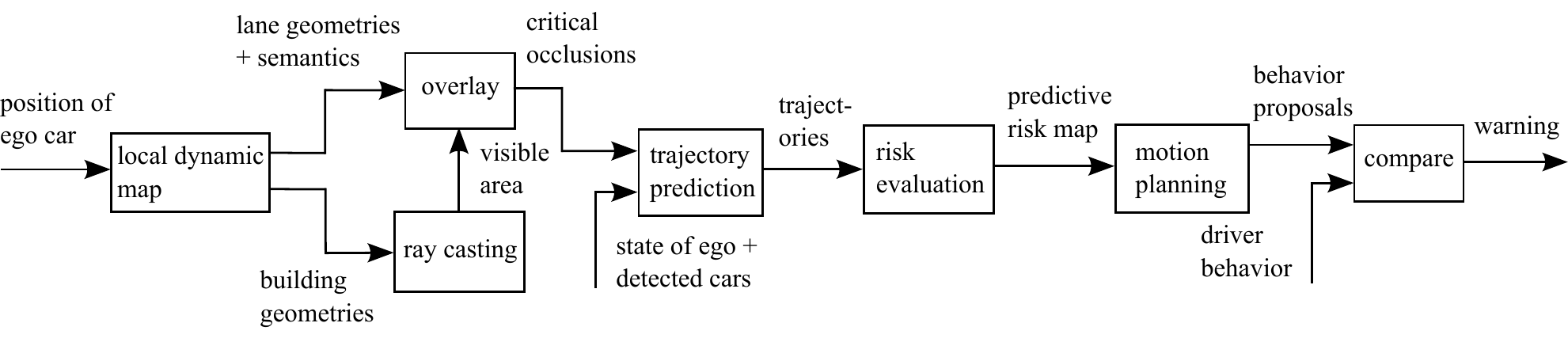}
      }
      \caption{Warning system for intersections of limited visibility by modeling virtual cars in occluded areas.}
      \label{fig:approach}
\end{figure*}

By looking at the behavior of cars with limited visibility area, one can infer the occluded road structure or presence of potential pedestrians. Consequently, an occupancy grid is created in \cite{afolabi2017}. 
As an alternative, the authors of \cite{toepfer2013} propose to divide detected lane structures into patches and to connect them as an undirected graphical model. Geometrically extrapolating the road out of the sensor range allows to infer information at occlusions. 
For representing map data, a suitable model is required. Bender et al. introduce the so-called lanelets which are atomic lane segments, defined by its left and right boundary polyline with the aim of storing topological and metrical infrastructure information \cite{Bender2014}. In \cite{Betaille2010a}, the road representation consists of basic curves such as straight lines, circles, and clothoids (spiral curves), which all share the same curvilinear 2D equation.
\cite{guzman2010} combines a local dynamic map with vehicle-to-vehicle (V2V) connectivity to localize vehicles behind buildings at intersections. 

Numerous approaches for calculating risk on longitudinal scenarios exist in literature, but risk at intersections is still an open research topic. The authors of \cite{armand2016} learn a typical velocity profile of vehicles approaching intersections with a Bayesian network. The result is a value that descibes the probability of the occurence of a dangerous situation while approaching another vehicle. \cite{lefevre2012} takes vehicles on intersections into account to retrieve an interaction probability. 
In \cite{beaucorps2017}, velocity profiles are clustered 
into several candidates and used to calculate the post encroachment time (PET) as a risk indicator. 
Lastly, \cite{kamijo2000} tackles accident detection at intersections with a hidden Markov model (HMM) and its features relative velocity, orientation and position of two objects.

Motion planning techniques allow to find trajectories in the driving space which are optimized towards driving duration with the restraint of being collision-free.
For this purpose, \cite{andersen2013} employs a state machine with model predictive control (MPC) for overcoming static obstacles while \cite{isele2017} trains a Deep Q network (DQN) to depart safely for left and right turning plus forward passing at unsignalized intersections. In a related but coarser approach \cite{quinlan2010}, the authors compare V2I intersection managers with stop sign or traffic light based intersections and show that V2I intersection managers can reduce the delay of vehicles. Simpler techniques are incorporating V2V for an improved traffic light control \cite{loos2011}.

In total, increasing work try to tackle limited visibility problems in inner-city scenarios due to low robustness in object detections of active sensors. Cameras cannot reliably detect other vehicles at night or in glaring images [6,7]. Lidar does not have a long range [8] and radar holds small opening angles. To overcome car detection issues, data-based solutions [16,17,21] lack the generalization to arbitrary dynamic scene evolutions and static road geometries. V2V or V2I
[15,22,23] are able to function theoretically in all cases, but create practical time synchronization and spatial alignment issues for all connected vehicles. Finally, previous model-based solutions [11,12,18,20] are often tailored occlusion heuristics and fail to handle real collision risks.

\subsection{System Overview} \label{sec:approach}
At intersections, the risk estimatoin is dominated by approaching traffic participants. Depending on traffic rules, the collision risk with each has to be estimated. 
However, numerous areas around inner-city intersections are difficult to access visually since they are occluded by buildings. 
Therefore, the occlusion risk cannot be derived directly. The ego driver has to assume potentially present but indetectable entities and their trajectories. 

Taking this into consideration, an example is the act of slowing the ego vehicle down when approaching an intersection of limited observability to be able to stop in a right-of-way situation.
Once the intersection is fully observable and a safe crossing is ensured, the ego car can retake speed and keep on driving. Our approach mimics this concept and its block diagram is shown in Fig. \ref{fig:approach}.

The basis forms the R-LDM, which is realized as a connected graph database storing road and building information.
Starting from the current position of the ego car, we first estimate the detectable area at the intersection. Therefore, we use a ray casting algorithm \cite{compg2008} on the basis of surrounding buildings which are used as occluding objects. In this process, the sensor range is reduced.
We overlay the visible area with map data containing the geometries
of relevant incoming lanes in order to determine occluded lane segments with potential risk sources. On these segments, virtual cars are modeled. Subsequently, we predict future spatio-temporal trajectories with constant velocity models for the virtual cars.

A variation of possible ego trajectories in combination with fixed trajectories of the other cars 
are evaluated in terms of collision risk which allows to build predictive risk maps \cite{Damerow2014Risk}. 
Exploiting the risk map, we calculate constant velocity, deceleration and acceleration trajectories and
select the ones with low-risk. 
By comparing the driver's current behavior with the alternatives, we achieve an ADAS that is able to warn the driver in cases in which approaching the intersection is considered to be dangerous. In addition, the most suitable choice can be communicated.

The remainder of this paper is structured as follows: in subsequent Sections \ref{sec:ldm}, \ref{sec:occrisk} and \ref{sec:warnsys}, individual components of the risk-based driver assistance are described in more detail. 

\setcounter{figure}{2}
\begin{figure*}[t!]
      \centering
      \includegraphics[width=1.0\linewidth]{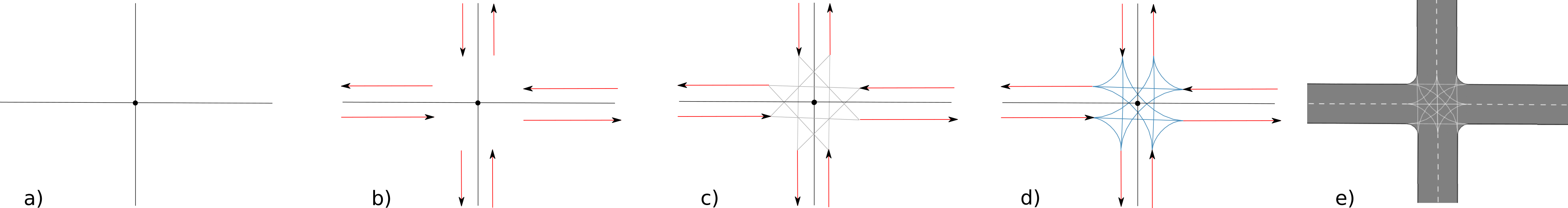}  
      \caption{Map data enhancement. a) Road centerlines. b) Calculation of lane centerlines. c) Inference of topological connections. d) Interpolation of junction centerlines. e) Full intersection geometry.}
      \label{fig:map_enhancement}
\end{figure*}
\setcounter{figure}{1}
\begin{figure}[h!]
      \centering
      	\includegraphics[width=0.55\linewidth]{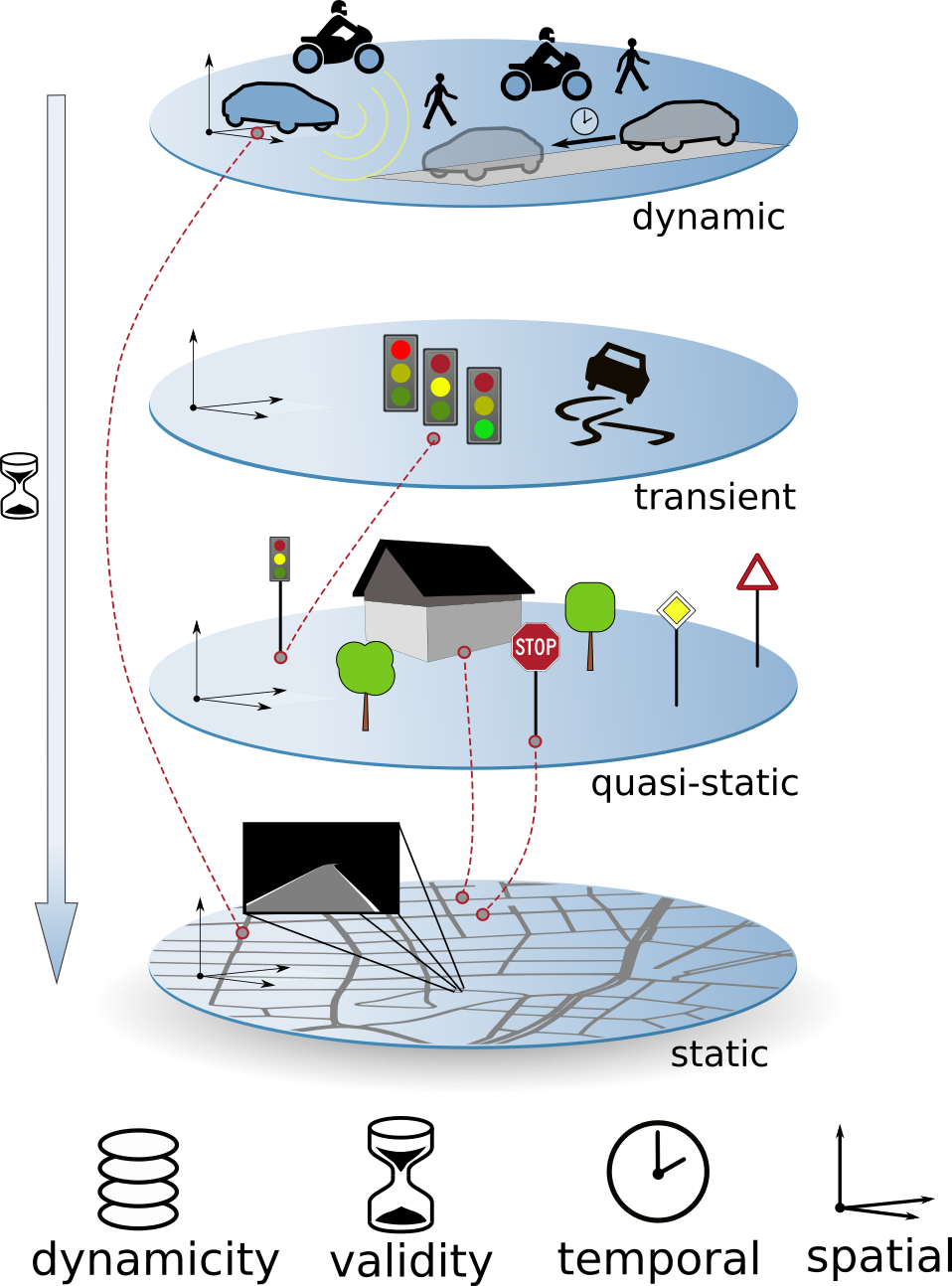}  
      \caption{Graph-based R-LDM concept.}
      \label{fig:ldm_concept}
\end{figure}
\setcounter{figure}{3}

\noindent In Section \ref{sec:application}, we validate the system on real world intersection scenarios. To conclude, Section \ref{sec:conclusion} shows research areas for further improvement.

\section{Relational Local Dynamic Map} \label{sec:ldm}

\subsection{Concept} 

Intelligent vehicles require knowledge about the ego position and about the position of other related traffic entities.
Such data ranges from  absolute GNSS coordinates over vehicle-relative lidar measurements to building information denoted in Cartesian coordinates. 
A solution to this task presents the local dynamic map, which is serving as a hub for receiving, integrating, storing, fusing, updating and predicting ADAS-relevant data. It allows to collect and manage environment information from different sources.
Similar to the SAFESPOT project \cite{Safespot2008} and its subsequent research, our R-LDM is realized as a 4 layer model, see  Fig.~\ref{fig:ldm_concept}.
On its first or lowest layer, static data is managed, mainly consisting of road infrastructure related information which is changing on a slow timescale.
The second layer handles quasi-static data which is changing on a scale of several days or longer. 
Examples are traffic rules expressed as traffic signs, roadside infrastructure, road constructions, buildings or trees. 
On the third layer, transient data is managed which is changing after some hours or faster. Whereas traffic lights themselves are assigned to the quasi-static layer, traffic light phases belong to the transient layer. 
Further examples of transient data are traffic density, congestion areas and temporal road changes like slippery road conditions. 
Last but not least, dynamic data is stored in the highest layer including the state of other traffic participants such as vehicles or pedestrians. 

\subsection{Data Enhancement} 

Yet unanswered is the question of the data source.
While the majority of entities from upper layers are sensed online or exchanged via communication protocols, entities from lower layers mostly rely on offline data. 
Especially geometrical and topological mapping information needs to be obtained and stored in advance.
Approaches range from recording new data while using extensive sensor equipment \cite{Levinson2007} to creating map data by parsing aerial satellite images as presented in \cite{Mattern2010} or \cite{Mattyus2015}.

In our research, we use data from publicly available crowdsourced map databases, more specifically from OpenStreetMap (OSM) \cite{osm}.
Collaborative mapping projects offer map data with the advantage of their data being editable and updated regularly.
OSM data  contains geometry data on road or half-road level. In this context, we define a half-road as being the sum of all lanes in one direction which allows us to express directionality. Data provided by OSM is sufficient for navigational purposes on a strategic level. 
However, lane-level information, such as intersection topology and connectivity, is mostly not provided explicitely.
Therefore, our idea is to enhance basic OpenStreetMap data as previously presented in \cite{Cao2016a} and \cite{Flade2017}.

Fig. \ref{fig:map_enhancement} illustrates the enhancement process.
Starting from polylines, obtained from OSM (Fig. \ref{fig:map_enhancement}~a), we infer lane segment centerlines. 
This is done by using the information about the number of lanes, provided as tag, in combination with an assumed lane width. 
Based on this lane segment information, we then infer possible connections (Fig. \ref{fig:map_enhancement}~c) between lane segments, using heuristics.
Now having respective junction information on topological level, the next step is to generate geometrical lane junction information. 
Here, we interpolate between two related lane segments using {B-Spline} curves (Fig.~\ref{fig:map_enhancement}~d). 
In last optional steps, left and right delimiter polylines, describing the actual lane shape, can be added (Fig. \ref{fig:map_enhancement}~e).
With regard to the later storage of the generated data, we want to highlight that a segment is always followed by a junction and vice versa. Furthermore, all junctions at a certain location form an intersection. 

As noted, the map enhancement is based on strong assumptions.
The assumed lane width highly influences the inferred lane-level geometry.
According to the research of Hall et al., lane widths of freeways and arterial roads globally range from 2.75~m to 4.0~m. 
Considering also minor or local roads, the width ranges from 2.25~m to 4.0~m \cite{Hall1998}.
Nevertheless, taking local guidelines for urban road design of the individual country into account, the maximum error based on a certain road type can be reduced. 
For the example of Germany, one can then assume a maximum lane width error of +-25~cm.
In order to improve the accuracy, it pays off to compare the automatically enhanced data with superimposed aerial images in order to perform visual consistency checks.

In addition to road infrastructure information, the second type of OSM-based information is data on buildings which is part of the second or quasi-static layer.
OSM provides the outline of buildings in forms of polylines as well as explicit height information in meters or implicit height information in form of the number of floors.
However, this information is not provided reliably.
Furthermore, it has to be noted that special shapes such as balconies are not considered in our research.

\subsection{Storage of Data}

As introduced in \cite{ldm2017}, 
our R-LDM is realized  as a native graph database, more specifically a labelled property graph model.
Regarding the static layer, we store geometrical and topological information on up to three different layers of detail, depending on the requirements.
The most detailed and lowest abstraction layer are lanes. 
With regard to multi-lane roads, all lanes pointing in the same direction form the so called half-roads, presenting the second layer.
A road, being the third layer, therefore consists of up to two half-roads which by themselves can consist of an arbitrary number of lanes. 

For each element of every layer, it is possible to store properties, ranging from current geometry information up to surface material information.
While properties can be chosen freely, it is recommended to always include information describing the shape, e.g. the center of a lane segment defined by a polyline or the outline of a building defined by a polygon.
All spatial references are denoted in geographic coordinates.
For junctions, it has proven useful to store the angle between the two segments that are connected by the respective junction. This is done by calculating the angle between the vector of the first two polyline points before and after the junction element.
Furthermore, we express priority rules by connecting two lane segments via an additional relation.

\subsection{Data Retrieval} 
Choosing a graph database pays off in the data retrieval step.
Since relationships  are first-class citizens of the graph data model, new entities can easily be connected to any other existing element from any layer which allows the embedding of entities into a context.
Especially when starting from a certain map entity, this interconnectivity enables convenient and quick information retrieval. 
Intuitively, a map entity such as a vehicle can be connected to another map entity such as a lane.
Correspondingly, we connect buildings to the associated half-road entity.
Since half-roads are defined by their directionality, we implicitely know whether a building is located on the left or right side of the road, just by quering for the connected half-road segment. 
In the case of static map elements, it is easily possible to query for neighbouring, following or preceding entities.
In the same way, it is possible to retrieve connected entities of a different abstraction layer, such as all lane segments assigned to a specific road segment.
This also allows us to intuitively obtain a so-called ``path horizon'', describing all possible paths a vehicle can follow based on its current position which is e.g. needed 
for inference of future paths and intentions. 

A second possibility for querying entities is to use their spatial reference.
This is especially useful if no starting node is given, when retrieving entities within a certain radius around a given position or when retrieving the closest entity of a certain type. 
In this case, we use Global Navigation Satellite Systems (GNSS) such as GPS, GLONASS or Galileo to obtain a coarse absolute position estimation. 
We use an R-Tree, embedded into the graph structure, for efficient indexing from metric coordinates (e.g. latitude / longitude) to graph elements. 
R-Tree data structures group objects by their location and store the minimum bounding rectangle of each group. 
Besides quick graph entry, an R-Tree allows for efficient retrieval of entities within a certain spatial radius around a given position or for the closest entity of a certain type, even if no explicit relations exist (yet). 
Such spatial reference then allows us, for example, to query for all buildings within a certain radius or to retrieve possible paths based on the current position.
In the latter case, it is important to keep potential localization inaccuracies or offsets between map data and GNSS data in mind.
One way to reduce the effect of localization errors is to perform consistency checks, e.g. by checking if the direction of travel according to GNSS fits the direction of the expected closest lane segment.

\section{Occlusion Risk} \label{sec:occrisk} 

\subsection{Estimation of Visible Area} \label{sec:visible_area}

\setcounter{figure}{3}

\begin{figure*}[t]
      \centering
      \framebox{\parbox{1.0\linewidth}{
      \centering
      \includegraphics[width=\linewidth]{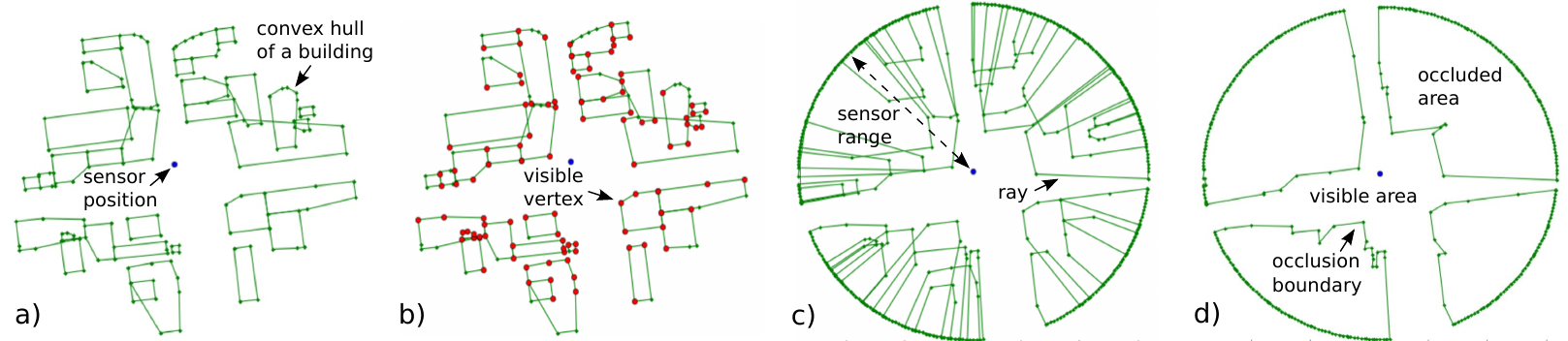}
      }}
      \caption{Occlusion properties of buildings. a) Buildings represented by their convex hulls. b) Filtering of visible vertices. c) Ray casting in sensor range. \newline d) Union of occlusion polygons.}
      \label{fig:convex_hull}
\end{figure*}

The estimation of sensory observability, here especially the estimation of the visible area at intersections, is crucial for the evaluation of risk from non-perceivable traffic scene entities.
This area is treated as a mathematical polygon in the field of computational geometry. Since the visibility is sought for the ego vehicle, it is categorized as a point visiblity polygon \cite{wikray}.   

In this work, we focus on occlusions caused by static buildings. 
To a certain degree of accuracy, we assume that the geometry of buildings is stored in the LDM.
In a preprocessing step, 
we query all buildings close to an upcoming intersection and 
represent each building by its convex hull. Labeled points inside the building description are thus erased. Fig. \ref{fig:convex_hull} a) depicts the output for a generic example where the sensor is positioned in the middle of an intersection. The number of relevant points is further reduced in Fig. \ref{fig:convex_hull} b) by filtering their visible vertices. The convex hulls are therefore transformed from Cartesian into polar coordinates. 

To find the visibility polygon, we start with the sensor's theoretical detection area. For simplicity, we assume a circle with a radius $r=\unit[50]{m}$ around the sensor position. Within this circle, the sensor provides reliable measurements. 
In a next step we use a ray casting algorithm, where we target only the visible vertices. As a result, we gain occlusion polygons (see Fig. \ref{fig:convex_hull} c). The occluded area represents the union of all occlusion polygons, whereby the 
visible area is found by geometrically subtracting it from the sensor circle. Fig. \ref{fig:convex_hull} d) shows the occlusion boundary line which separates the visible area from the occluded one. 

\subsection{Modeling of Virtual Cars} \label{sec:entity_modelling}

The goal of the system is to assess the upcoming risk for evaluating the current ego entity's behavior and allowing the planning of a risk-aversive behavior. This implies the prediction of possible future scene evolutions. Besides entities captured by sensors, it is 
necessary to consider entities that cannot be sensed but which potentially cause high risks for the ego vehicle. These are entities 
which approach the same intersection as the ego vehicle, but which are occluded by nearby buildings.
Consequently, the system has to estimate possible positions, where such non-observable critical entities may be located and it has to predict their future behavior. 


\begin{figure}[b!]
      \centering
      \framebox{\parbox{0.95\linewidth}{
      \centering
      \includegraphics[width=\linewidth]{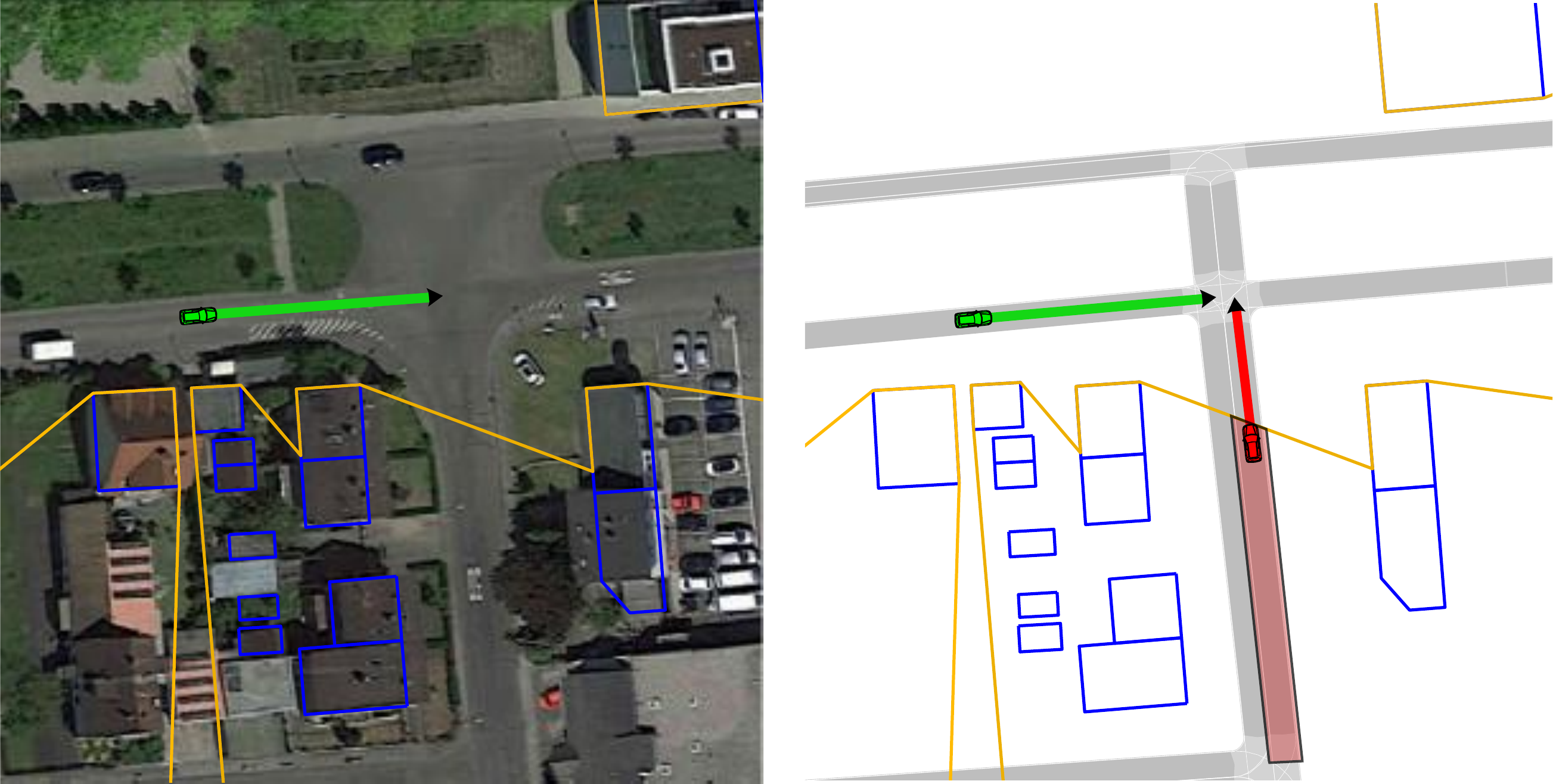}
      }}
      \caption{Example of a partially visible intersection. Left: Satellite view \cite{GM} and visibility area. Right: Positioning of virtual car.} 
      \label{fig:visible_vehicle_locations}
\end{figure}

Topologic road information in the R-LDM defines how incoming lanes are connected to outgoing lanes at intersections. 
We select those incoming lanes which have right-of-way priority over the ego vehicle's current lane. Then, we overlay their geometry with the visible area gained in Section \ref{sec:visible_area} to achieve an estimate of relevant occluded lane segments. 

A virtual car may be situated anywhere on each segment with highly 
uncertain behavior. 
We introduce only one virtual car at the occluded position closest to the intersection for every segment.
The car is assumed to drive with constant velocity along the lane's centerline (i.e. $\unit[40]{km/h}$ for urban traffic). 
Once its position is at the most critical location, right in the middle of the intersection, we assume a sudden stop in the trajectory prediction. 
This represents a worst-case-like behavior for the ego driver and enables a computationally inexpensive way to reproduce different position and velocity profiles of the virtual car.

Fig. \ref{fig:visible_vehicle_locations} illustrates a green ego car approaching an intersection. 
The visible area (orange) is limited due to occlusion caused by buildings (blue) and a virtual car (red) is located on the relevant lane (light red) at the boundary of the visible area 
with a longitudinal velocity profile pointing to the intersection center.

\subsection{Collision Risk Evaluation} \label{sec:risk_estimation}
In general, risk is the expectation value of the cost related to critical future events \cite{riskdef}. The evaluation of risk includes a prediction of those critical events as well as an estimation of the damage in case a related event occurs.
Future risk\footnote{To be precise, this is the risk \emph{density} over time.} is defined as the cost expectation value
\beq\label{eq:basic_concepts:dynamic_scenes:EqRisk}
r(t+s,\xvec_t)=\int c_{t+s} \, P(c_{t+s}|\xvec_t) \, dc_{t+s},
\eeq
where $P(c_{t+s}|\xvec_t)$ is the probability of a damage $c_{t+s}$, happening at future time $t+s$, for the known states $\xvec_t$ of the current scene.

Since the scene may be composed of several entities (e.g.~different traffic participants) with state vectors $\xvec_t^i$ (ego car state $\xvec_t^0$), we write $\xvec_t:=\{\xvec_t^0,\xvec_t^1,...,\xvec_t^n\}$. Using discrete time step indices $t,\, t+1,\, ...,\, t+s$ (time step size $\Delta t$), state vector sequences are additionally introduced as
\beq
\label{eq:basic_concepts:dynamic_scenes:trajectories}
\xvec_{t:t+s}:=\{\xvec_t,...,\xvec_{t+s}\}
,\eeq
which describe the state of the scene from $t$ (now) until a time $t+s$ ($s$ into the future).

\begin{figure}[b!]
      \centering
\resizebox{1.0\linewidth}{!}{
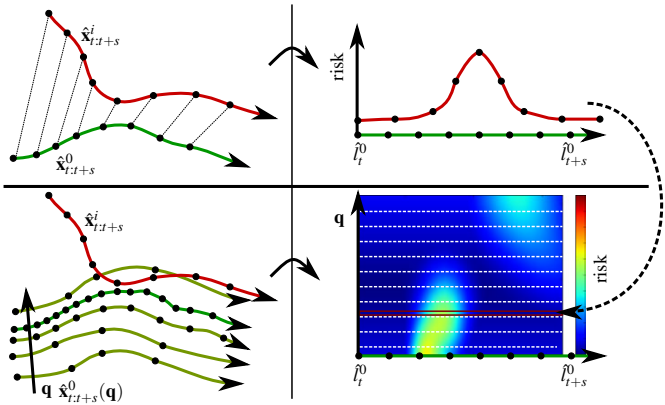}
      \caption{Generation of a predictive risk map. Top: Use of risk model. Bottom: Variation of ego car's predicted velocities.} 
      \label{riskmap_concept}
\end{figure}

According to \cite{Eggert2014Risk}, by assuming risks being caused by rare critical events and incorporating approximations on the probabilistic state vector sequence ${\xvec}_{t:t+s}$ by a prototypically predicted state vector sequence $\hat{\xvec}_{t:t+s}$ and a deterministic damage calculation $\hat{c}_{t+s}$, the general expression for future risk is proposed as
\bea{
\label{eq:basic_concepts:dynamic_scenes:situation_dependent_risk}
r(t+s,\xvec_t)\sim &\sum_{e_{t+s}}
\hat{c}_{t+s}(e_{t+s},\hat{\xvec}_{t+s}(\xvec_t)) \nonumber \\
&\qquad \cdot P(e_{t+s}|\hat{\xvec}_{t:t+s}(\xvec_t),s).
}
\noindent $P(e_{t+s}|\hat{\xvec}_{t:t+s}(\xvec_t),s)$ describes the event triggering probability with the discrete variable $e_{t+s}$ describing a certain critical event at future time $t + s$,
such as car-to-car, car-to-pedestrian or car-to-infrastructure collisions.

In order to advance from a general definition of risk to an actual risk measure for a given prototypically predicted state
vector sequence, 1) the deterministic damage calculation using a damage
approximation model and 2) the probability of the future event happening
at time $t + s$ have to be modeled.
	
The event probability $P(e_{t+s})$ of a single event $e_{t+s}$ can be expressed as the combination of the instantaneous event rate $\tau_{e_{t+s}}^{-1}$ (assuming that the entity ``survives'' until future time $t+s$) and the probability that the entities actually survive
\bea{
\label {eq:basic_concepts:general_risk_model:event_probability}
&P(e_{t+s}|\hat{\xvec}_{t:t+s}(\xvec_t),s) \nonumber \\
&\ \ \ =\tau_{e_{t+s}}^{-1}(\hat{\xvec}_{t+s}(\xvec_t),s)  S(\hat{\xvec}_{t:t+s}(\xvec_t), s)\delta t .
 }
Here, the so-called inhomogeneous survival function $S$ takes into account that the observed entities possibly have already been involved in another critical event, e.g. a collision, before it could actually be involved in the considered event
\bea{
\label{eq:basic_concepts:general_risk_model:survival_funktion_ext}
&S(\hat{\xvec}_{t:t+s}(\xvec_t), s)  \nonumber \\
&\qquad=\exp\{-\int_0^s \tau_0^{-1} + \tau^{-1}(\hat{\xvec}_{t+s'}(\xvec_t),s') \, ds'\},
}

\noindent with the total event rate, combining all possible event rates
\bea{
\label{eq:basic_concepts:general_risk_model:total_event_rate}
\tau^{-1}(\hat{\xvec}_{t+s}(\xvec_t),s)=\sum_{e_{t+s}}  \tau_{e_{t+s}}^{-1}(\hat{\xvec}_{t+s}(\xvec_t),s)
.}

In \cite{damerow2018}  different event rates have been modeled. Exemplarily, for car-to-car collision risks, an appropriate model for the instantaneous event rate depends on the distance between two traffic participants such that the instantaneous event rate is large for small distances and
decreases with an increase in predicted distance
\bea{
\label {eq:basic_concepts:general_risk_model:eq4a}
&\tau_d^{-1}(\hat{\xvec}_{t+s}(\xvec_t),s) \nonumber \\
&\quad=\tau_{d,0}^{-1} \exp\{-\beta_d(s) \cdot\text{max}(\hat d_{t+s}(\xvec_t)-d_{min},0)\}
,}
where $\hat d_{t+s}(\xvec_t)$ is the predicted distance between the ego car and another traffic participant and $\beta_d(s)$ their spatial position uncertainty which grows with the predicted time $s$. The parameter $d_{min}$ stands for the minimal allowed distance corresponding to a physical overlap.

In Eq. \eqref{eq:basic_concepts:dynamic_scenes:situation_dependent_risk}, the expected cost/damage is modeled deterministically.
For car-to-car collision risks, an appropriate approximation of damage in case a collision occurs is modeled by the energy transfer between colliding entities. As a result, a 2D inelastic collision model is considered (more accurate damage models can be applied in similar ways), such that
\bea{
\label{eq:basic_concepts:risk:general_risk_model:collision_damage}
&\hat c_{t+s}(e_{t+s}, \hat \xvec_{t+s}(\xvec_t,h_t)) \nonumber \\
&\ \ \ \sim w_c \cdot \frac{1}{2}\frac{m_0 m_i}{m_0+m_i} || \hat v^0_{t+s}(\xvec_t,h_t) - \hat v^i_{t+s}(\xvec_t,h_t)||^2
,}
where $m_0$, $m_i$ are the masses and $\hat v^0_{t+s}(\xvec_t)$, $\hat v^i_{t+s}(\xvec_t)$ define the vectorial velocity components of the ego- and another entity involved in the collision risk estimation while $w_c$ is a weighting factor. The velocity components can be derived from the prototypically predicted state vector $\hat\xvec_{t+s}(\xvec_t)$, which relies on the current states of the scene $\xvec_t$.

The risk model \eqref{eq:basic_concepts:dynamic_scenes:situation_dependent_risk} is used to build predictive risk maps, as shown in Fig. \ref{riskmap_concept}. In the process, we do not only calculate the risk along a defined path $\hat l^0$ for one 
predicted trajectory of the ego car $\hat\xvec^{0}$ with respect to the other car's predicted trajectory $\hat\xvec^{i}$. Instead, we create a set of ego car trajectories
$\hat\xvec^{0}(\textbf{q})$ defined by variation parameters $\textbf{q}$ and evaluate the risk for each trajectory. When we use the predicted ego car velocity $\hat v_0$ as $\textbf{q}$,
a predictive risk map can be composed. It indicates how risky a chosen ego velocity will be for the predicted time $t+s$.

\section{Driver Assistance} \label{sec:warnsys} 
\subsection{Planning of Risk-Aversive Behaviors} \label{sec:behavior_planning}

\begin{figure}[t]
      \centering
\resizebox{\linewidth}{!}{
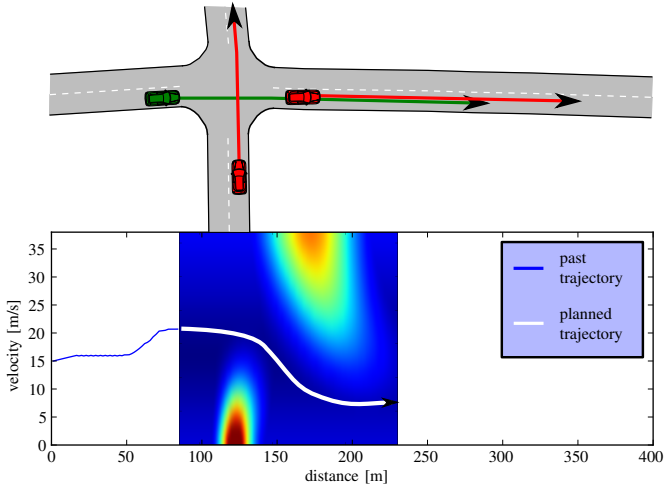}
      \caption{Behavior planning using risk maps. Top: Intersection scenario with two other cars. Bottom: Risk map with planned trajectory.}
      \label{fig:riskmap_planning}
\end{figure}

Risk maps allow the evaluation of future behavior alternatives for the ego vehicle. 
A modified, globally optimizing version of the rapidly-exploring random tree search algorithm has been used in \cite{Damerow2015RRT} to plan the most suitable velocity profile 
as a ``path'' through the map, minimizing risk and maximizing utility considerations.
Fig. \ref{fig:riskmap_planning} displays the outcome of the algorithm for an intersection scenario with two other traffic entities. The future cost-optimized trajectory
swerves around the two corresponding hot spots in the risk map.

To fulfill computation time constraints, we use a planning algorithm of reduced complexity by 
only considering the three behavior alternatives: 
1) keep on driving with constant velocity, 
2) brake with constant deceleration to be able to stop at the stop line of the intersection and 3) accelerate with constant acceleration 
to safely pass in front of the non-detectable approaching vehicle.

In a first step, the three behavior alternatives have to be calculated. Each is represented by a deceleration/acceleration value.
For driving with constant velocity, we set 
\bea{
a_{\text{const}}= \unit[0] {m}/\text{s}^{2}
\label{eq:a_const}
.}

The deceleration that is needed to stop at the stop line is calculated according to
\bea{
a_{\text{stop}} = -\frac{v_0^2}{2~d_{\text{sl}}} 
\label{eq:a_stop}
,}
where $v_0$ is the current ego car velocity and $d_{\text{sl}}$ the distance to the intersection entry point or stop line.
The distance $d_{\text{sl}}$ is derived from our map data introduced in Section \ref{sec:ldm}.  

For passing in front of potentially approaching vehicles, we use the risk map to estimate the target velocity $v_{\text{trg}}$ that has to be reached at the intersection 
to pass with a certain low risk value. The required acceleration is then 
\bea{
a_{\text{acc}} = \frac{v_{\text{trg}}^2-v_0^2}{2~d_{\text{cp}}} 
,\label{eq:a_acc}
}
with $d_{\text{cp}}$ describing the distance to the expected crossing point on the intersection between the ego and other vehicle's path.

In a second step, we evaluate the three behavior alternatives $a_{\text{const}}$, $a_{\text{stop}}$ and $a_{\text{acc}}$ within the risk map. 
Hereby, not only the collision risk from virtual vehicles, but also from detected vehicles by the ego vehicle's sensors are 
depicted in the risk map and have priority over occlusion risks.
In case the resulting risk value exceeds a threshold, the behavior alternative is neglected and not considered as a suitable action.

In \cite{Damerow2015Situation}, a situation classification is used as a preprocessing step prior to the behavior planning in order to consider different possible interactions of the traffic participants. 
The proposed approach for occlusion risk can seamlessly be incorporated into this full behavior planning framework.

\subsection{Driver Warning and Behavior Suggestion} \label{sec:warning_system}

\begin{figure}[t]
      \centering
      \resizebox{\linewidth}{!}{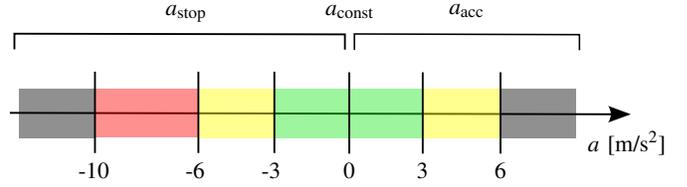}
      \caption{Levels of necessary braking or accelerating.}
      \label{fig:acc_dec_categories}
\end{figure}

We suppose an ADAS functionality that only steps in, when the currently performed behavior of the human driver is critical. 
For that purpose, we categorize the planned risk-aversive behaviors $a_{\text{const}}$, $a_{\text{stop}}$ and $a_{\text{acc}}$ in four levels of intervention according to Fig. \ref{fig:acc_dec_categories}: 
comfortable (green), heavy (yellow), emergency (red) and non-reachable (gray) \cite{burg2009verkehr}.

It can be seen that $a_{\text{const}}$ lies in the comfortable area.
However, the values of $a_{\text{stop}}$ and $a_{\text{acc}}$ can vary depending on the current state of the risk map (respectively the intersection and occlusion geometry) and 
thus reach different levels of intervention. We define that if both 
$a_{\text{stop}}$ and $a_{\text{acc}}$ have left the comfortable region, the driver behavior is seen as critical.
If those behaviors were applied, the car would reach its physical limits.
If $a_{\text{stop}}$ or $a_{\text{acc}}$ are in the non-reachable area, they have to be disregarded because it is not possible to execute them.

In compliance with the definition of driver criticality, we display a warning if $a_{\text{const}}=\unit[0]{m/s^2}$ is not safe (not among the proposed actions) and the alternative stopping and accelerating trajectories are outside of the comfortable region
\bea{
a_{\text{stop}}~\leq \unit[-3]{m/s^2} ~~~ \land ~~~ a_{\text{acc}}~\geq \unit[3]{m/s^2}. 
}
At this point, the behavior option with the lowest level of intervention can be provided to the driver.

On top of this, it can be considered to automatically activate a control mode if the driver is not responding appropriately.
As shown in Fig. \ref{fig:acc_dec_categories}, we do not consider  $a_{\text{acc}}$ as an appropriate emergency action.
Nevertheless, if the best possible action (lowest level of intervention) is $a_{\text{stop}}$ in the emergency area
\bea{
\unit[-10]{m/s^2}  \leq ~&a_{\text{stop}}~\leq \unit[-6]{m/s^2}
,}
the system could intensify the warning at first and initiate an emergency brake after waiting a predefined period of time.

\section{Simulation Results}
\label{sec:application}

We applied the proposed ADAS functionality for approaching intersections which are hard to access by the vehicle's on-board sensors to real world scenarios taken from the KITTI dataset
\cite{geiger2013kitti}. 
In general, the behavior of the human drivers in this dataset can be described as safe. Traffic rules and occlusions in urban environments were taken into account adequately. 
To evaluate driver behavior that can be considered as critical, we chose scenarios where the ego vehicle is actually on a priority road. 
However, we postulate that there is no priority information in the simulation. This results in a behavior, where the driver crosses intersections of limited visibility while neglecting 
potentially approaching but not detectable vehicles.\footnote{In the scenario with safe human behavior, we can thus exploit the taken velocity course of the test driver as a baseline and compare it with the planned velocities of our ADAS. The scenario with critical human behavior is only dangerous in our differently postulated resimulation. Nonetheless, this reflects real situations with occlusion-unaware traffic participants.} 


\subsection{Statistics of Visibility Area}
Depending on the visibility percentage of the intersection and the ego car's position and velocity, occlusion risk may be present. However, intersections and surrounding buildings alignment vary strongly. For demonstration purposes, the visibility estimation algorithm was tested on a 7:35 minutes long run with 40 intersections. We then examined the road visibility as the quotient of the visible and the total road area within a radius of $r= \unit[50]{m}$. The current lane of the ego car is excluded in this case.

\begin{figure}[t]
      \centering
      \parbox{1.0\linewidth}{
      \centering
      \includegraphics[trim={0 0cm 0 0.9cm},clip,width=\linewidth]{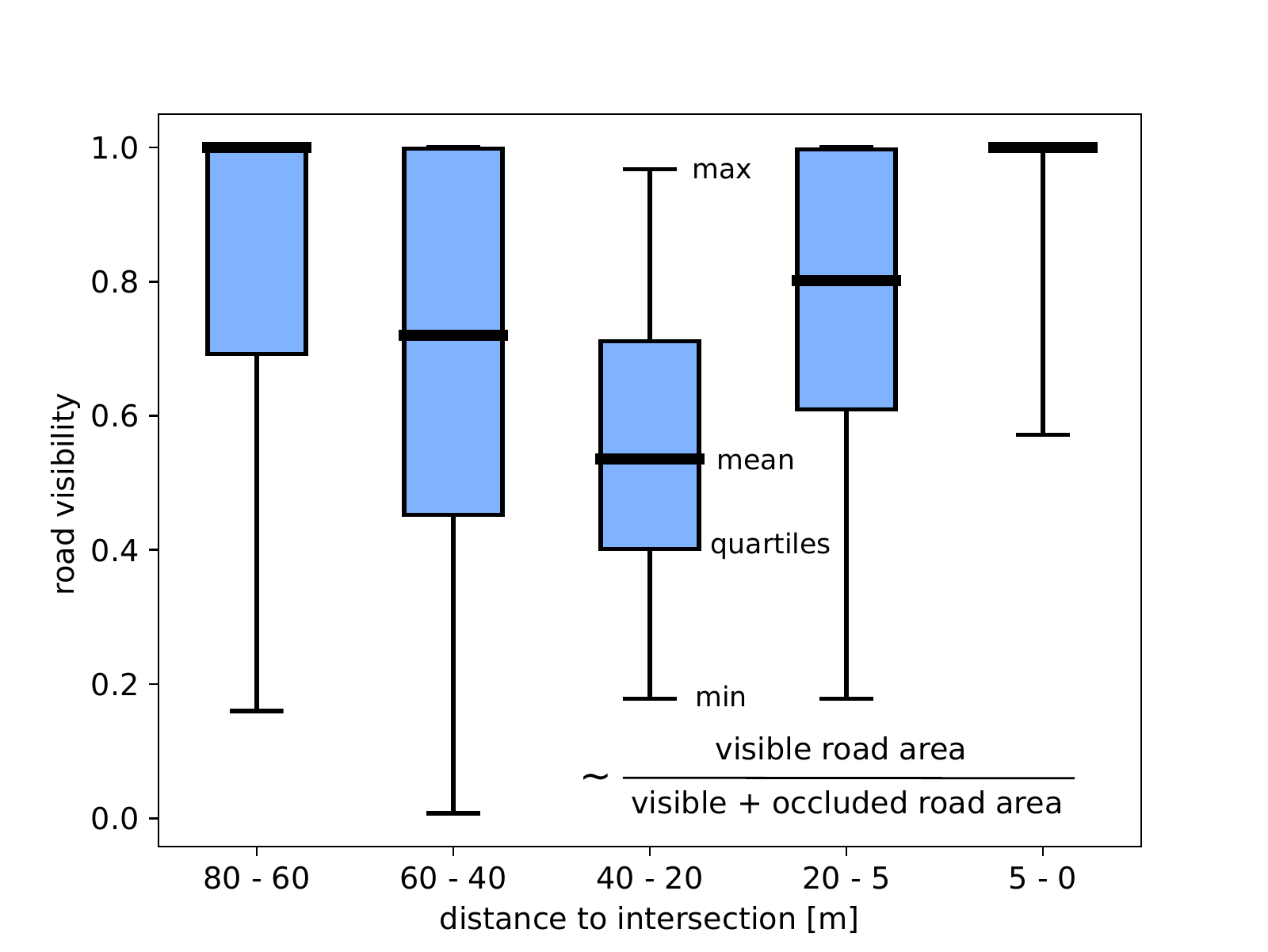}
      }
      \caption{Road visibility spread of different intersections.}
      \label{fig:visib}
\end{figure}

Fig. \ref{fig:visib} shows the outcome with a box plot. It consists of mean $\mu$ (middle stroke), first and third quartiles $\pm0.6745 \sigma$ (box) as well as min and max values (end of whiskers) for decreasing distance intervals to the intersections. While $\mu$ is 1 at $80 - \unit[60]{m}$ before, it declines to 0.5 in $40 - \unit[20]{m}$ and rises back to 1 for $5 - \unit[0]{m}$. 

The variance $\sigma$ is high for positions far away from the intersection ($80 - \unit[5]{m}$). On the one hand, there are intersections in which only few buildings are nearby. In this context, the road visibility stays approximately at the max value 1. On the 

\noindent other hand, there are intersections with high building coverage, leading to min values of 0 to 0.2. The box heights amount to 0.25 within $40 - \unit[20]{m}$, 0.3 inside $80 - \unit[60]{m}$ plus $20 - \unit[5]{m}$ and 0.5 at $60 - \unit[40]{m}$. Close to the intersection point at $5 - \unit[0]{m}$, $\sigma$ is small and the minimal road visibility is approximately 0.6.

Occlusion risk is in inverse proportion to the visibility and thus appears only before crossing the intersection. 
When the ego car is in the middle of the intersection, the risk disappears. To minimize the risk, the driver should react accordingly when approaching the intersection. 
To summarize, it can be said that occlusions play a dominant role in inner-city scenarios. On average, the worst-case visibility reaches down to $50\%$.

\subsection{Scenario with Safe Human Behavior}

\begin{figure}[H]
      \centering
      \framebox{\parbox{0.95\linewidth}{
      \centering
      \includegraphics[trim={0 2.9cm 0 3cm},clip,width=\linewidth]{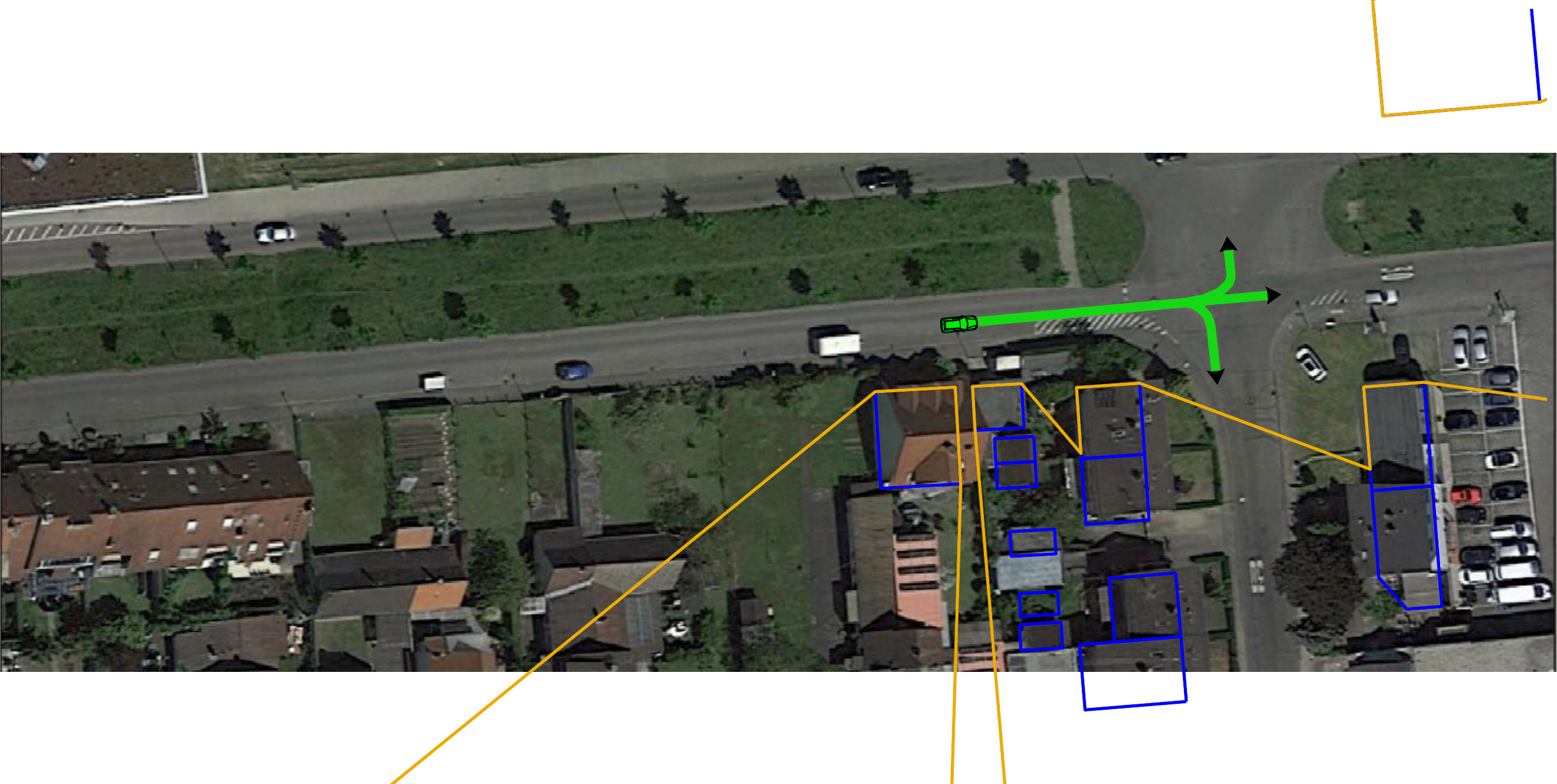}
      \includegraphics[trim={0 2.9cm 0 3cm},clip,width=\linewidth]{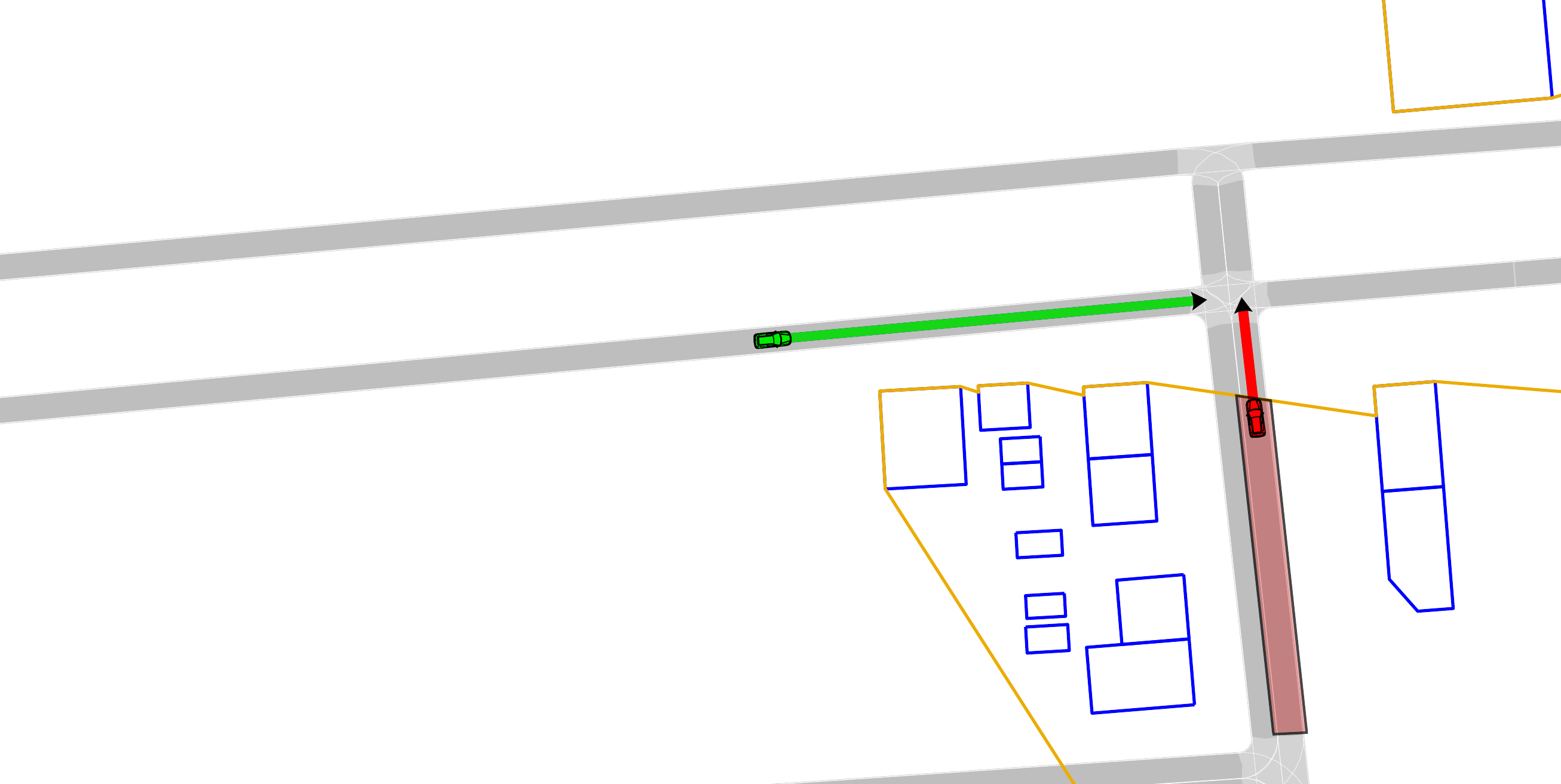}
      \includegraphics[trim={0.35cm 0cm 3cm 0cm},clip,width=0.97\linewidth]{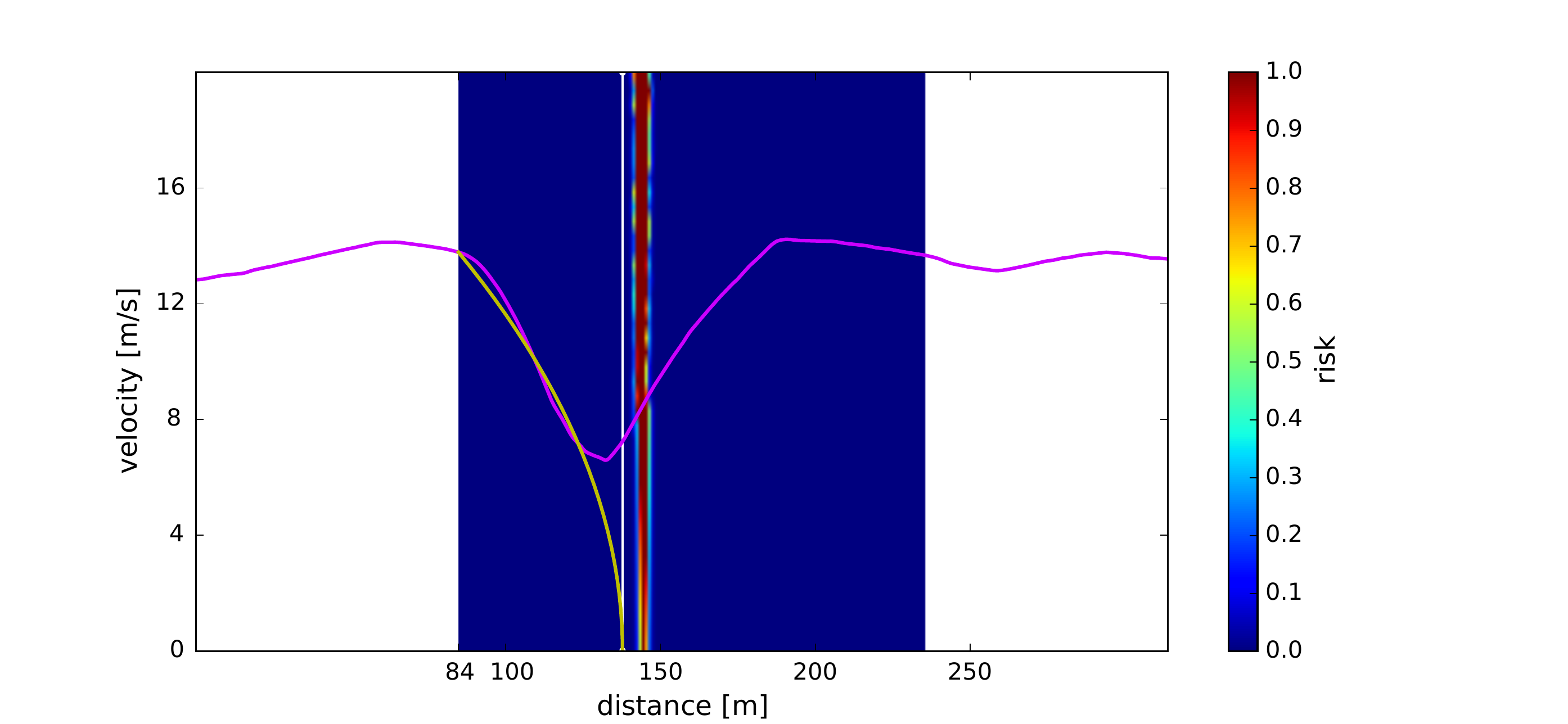}
      \llap{\includegraphics[trim={0.35cm 0cm 3cm 0cm},clip,width=0.97\linewidth]{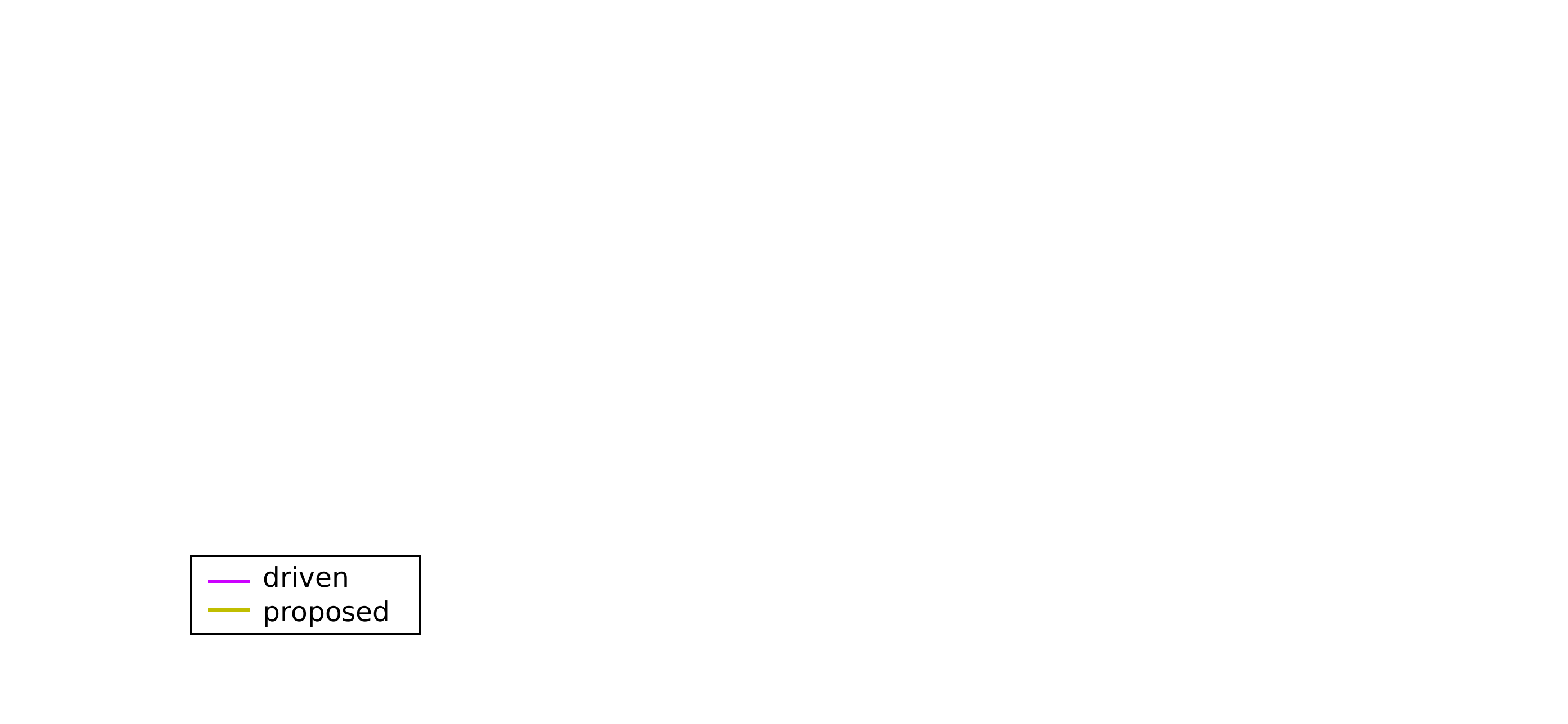}}    
      \llap{\includegraphics[trim={0.35cm 0cm 3cm 0cm},clip,width=0.97\linewidth]{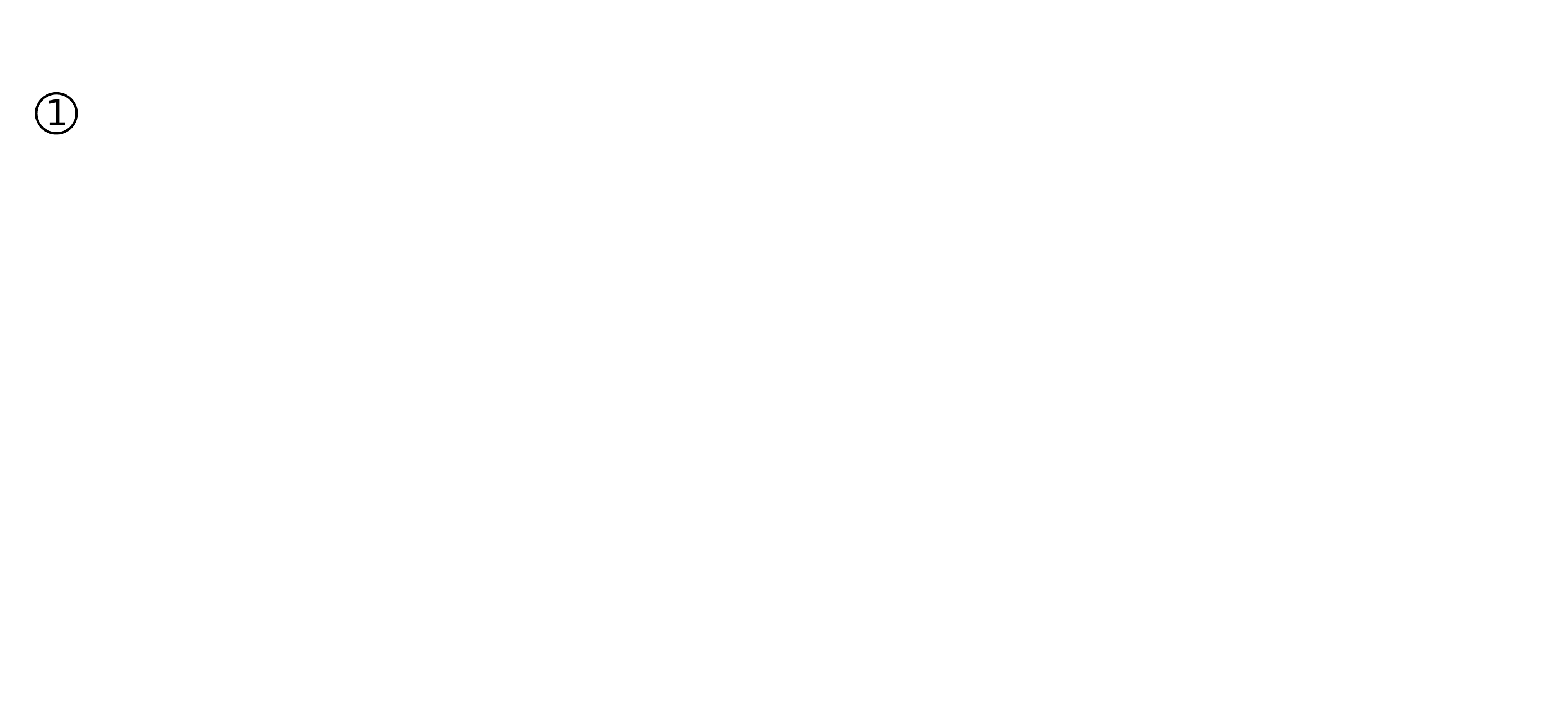}}    
      \includegraphics[trim={0 2.9cm 0 3cm},clip,width=\linewidth]{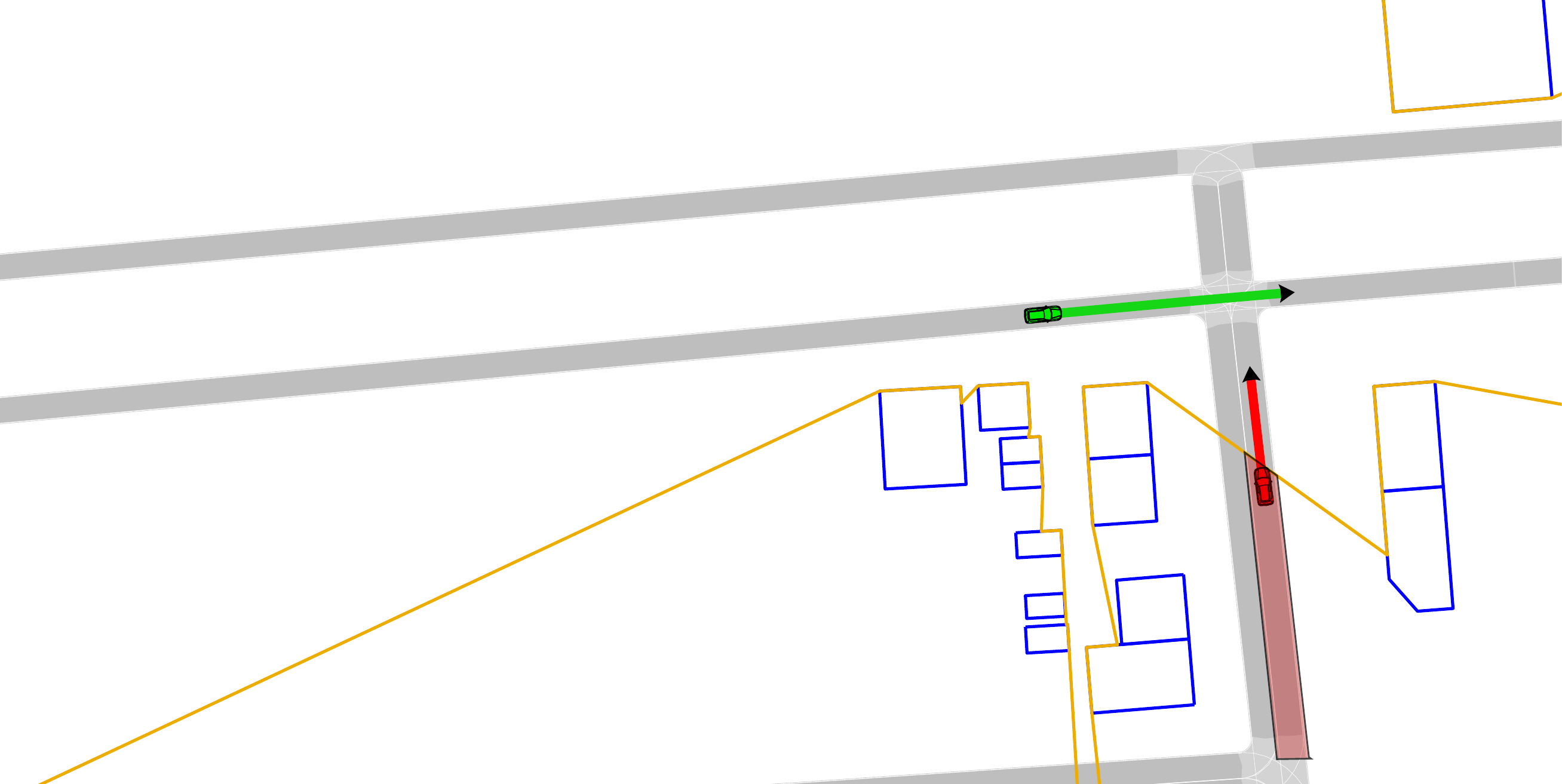}
      \includegraphics[trim={0.7cm 0cm 3cm 0cm},clip,width=0.97\linewidth]{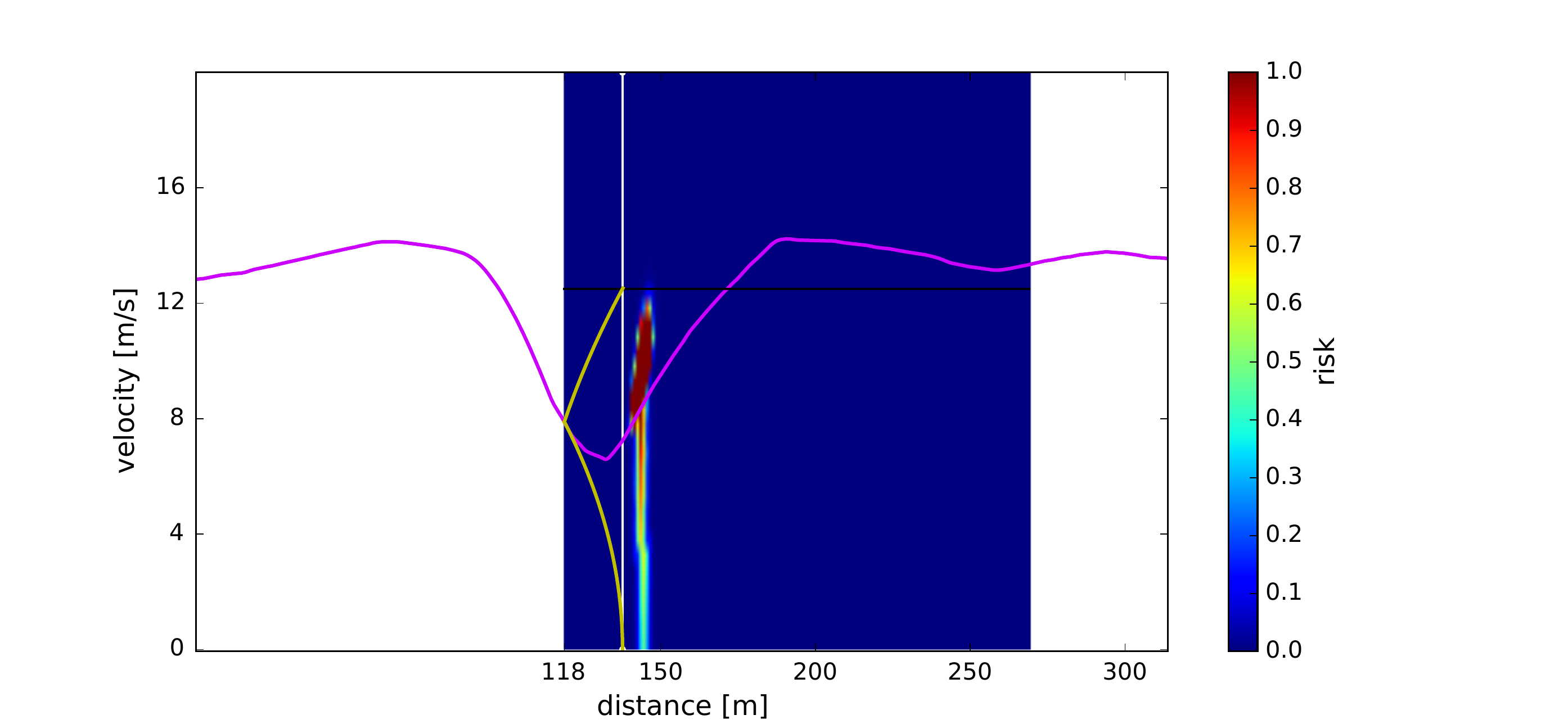}
      \llap{\includegraphics[trim={0.7cm 0cm 3cm 0cm},clip,width=0.97\linewidth]{images/fig10_4.pdf}}  
      \llap{\includegraphics[trim={0.35cm 0cm 3cm 0cm},clip,width=0.97\linewidth]{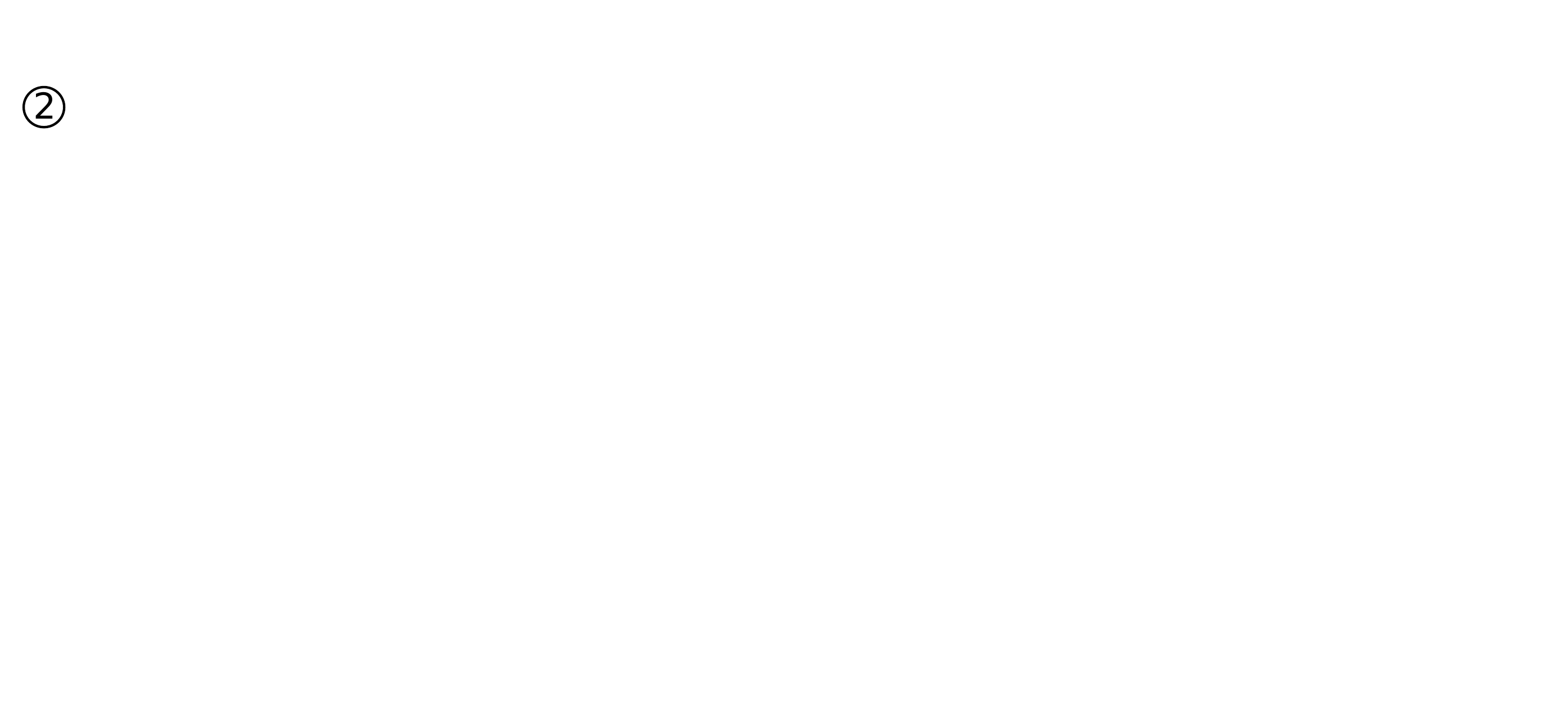}}    
      \includegraphics[trim={0 2cm 0 3cm},clip,width=\linewidth]{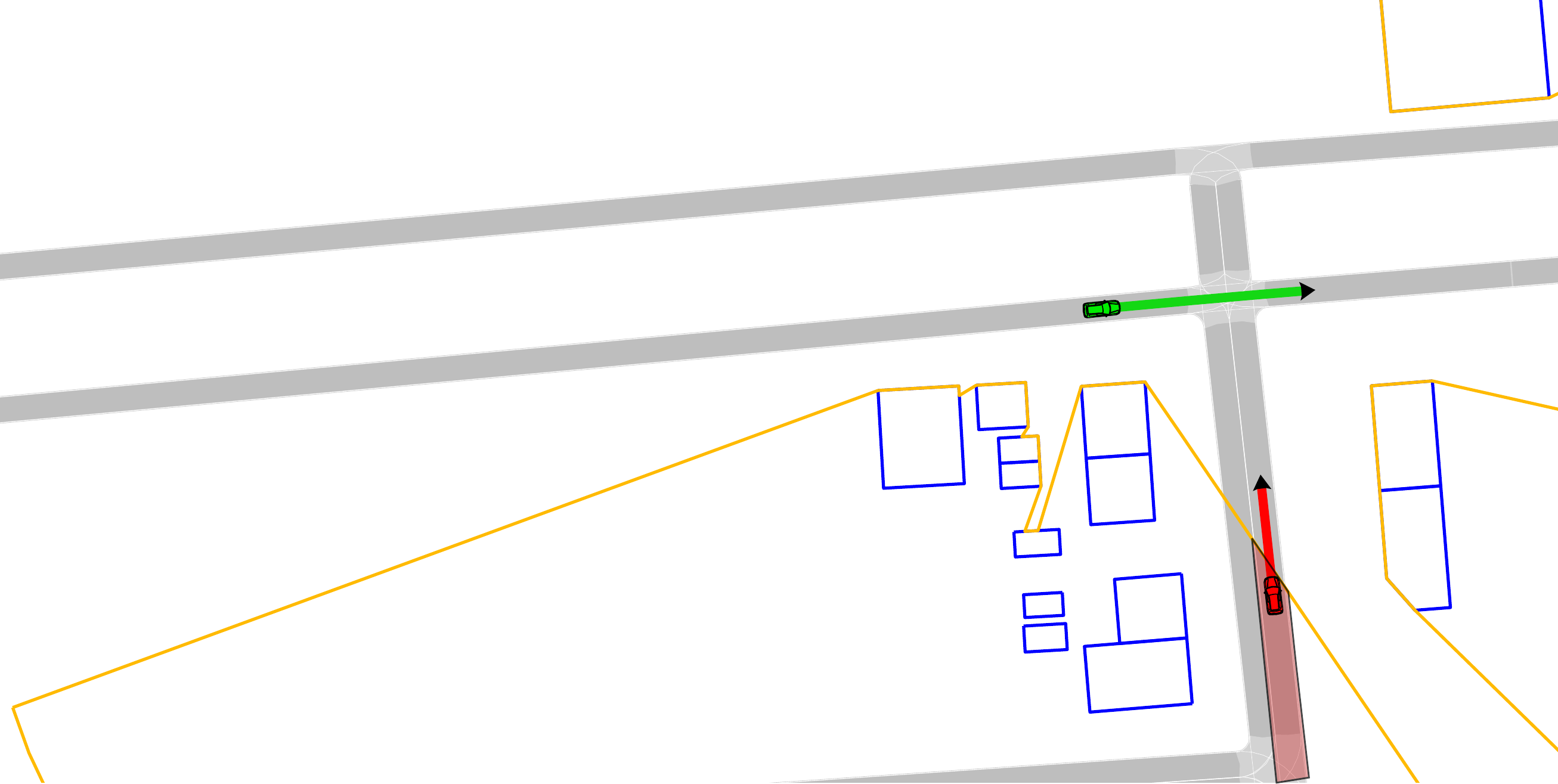}
      \includegraphics[trim={0.9cm 0cm 3cm 0cm},clip,width=0.97\linewidth]{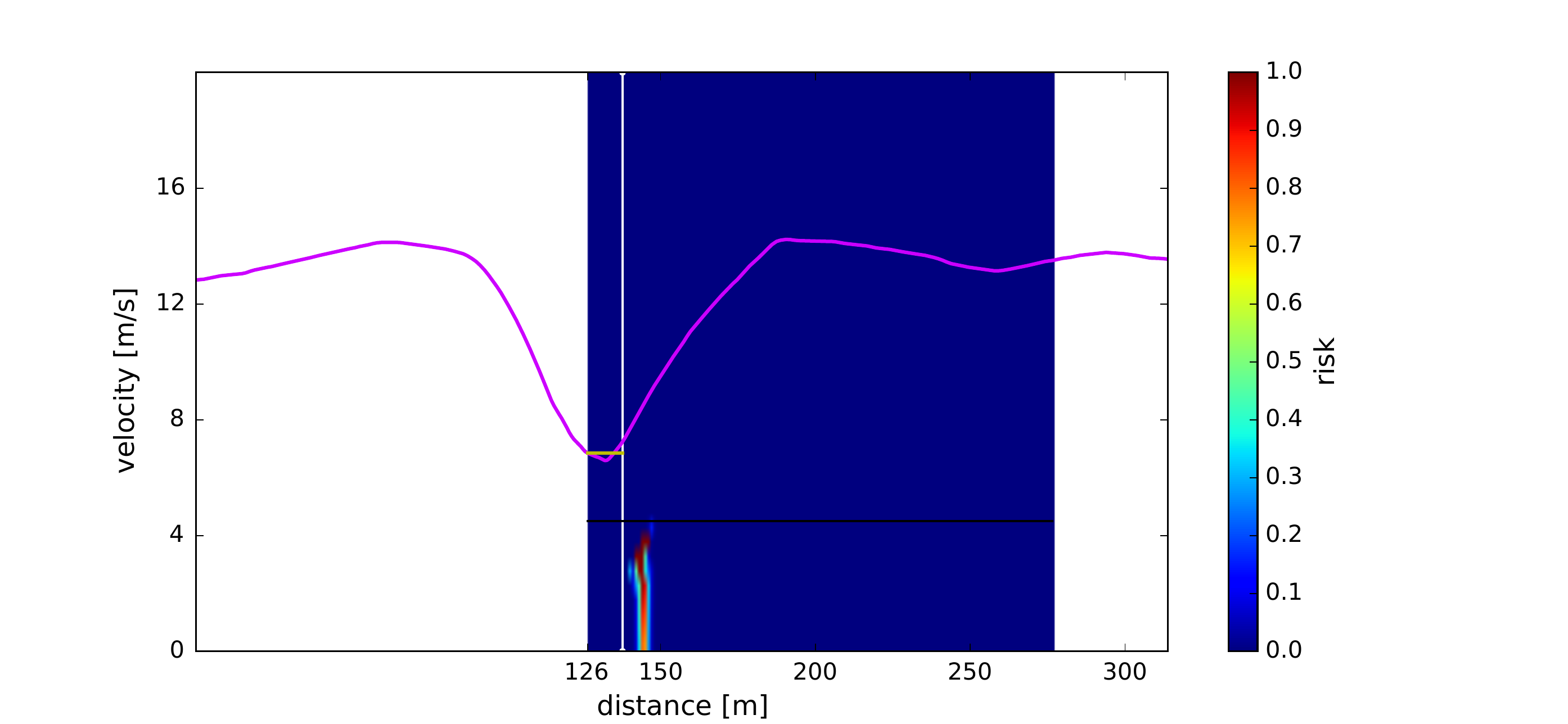}
      \llap{\includegraphics[trim={0.9cm 0cm 3cm 0cm},clip,width=0.97\linewidth]{images/fig10_4.pdf}}  
      \llap{\includegraphics[trim={0.35cm 0cm 3cm 0cm},clip,width=0.97\linewidth]{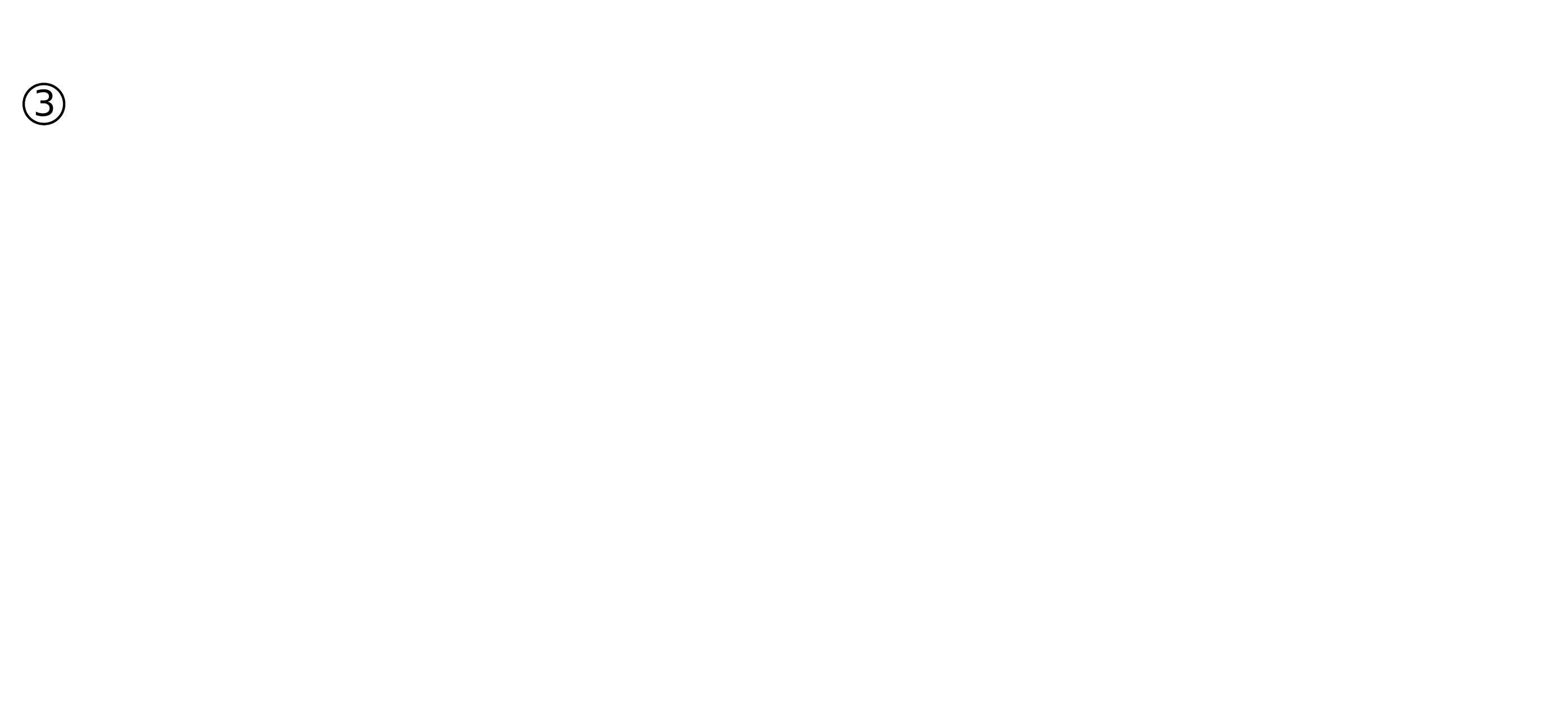}}    
      }}
      \caption{Evaluation of intersection assistance for limited visibility applied to a scenario with correct human behavior. Top: Satellite view and visibility area. Bottom: Simulation view and risk maps for the traveled distances 1) $d=\unit[84] {m}$, 2) $d=\unit[118] {m}$ and 3) $d=\unit[126] {m}$.}
      \label{fig:results_correct}
\end{figure}

Fig. \ref{fig:results_correct} outlines a satellite view of the ego vehicle approaching a partially visible intersection with buildings on the right limiting the field of view.
The modeling of the virtual car and the corresponding risk maps are shown for three consecutive time steps represented with the traveled distance $d=[84,118,126] \unit[]{m}$. 

Besides depicting risks from modeled virtual cars at occlusions, the plot also indicates the actually driven velocity profile of the human driver (purple line), 
the velocity profiles of the proposed behaviors $a_{\text{const}}$, $a_{\text{stop}}$ and $a_{\text{acc}}$ (yellow lines), 
the position of the stop line at the intersection $d_{\text{sl}}$ (white vertical line) as well as the velocity that is necessary at the collision

\begin{figure}[H]
      \centering	
      \includegraphics[trim={0cm 0cm 0cm 1.3cm},clip,width=0.97\linewidth]{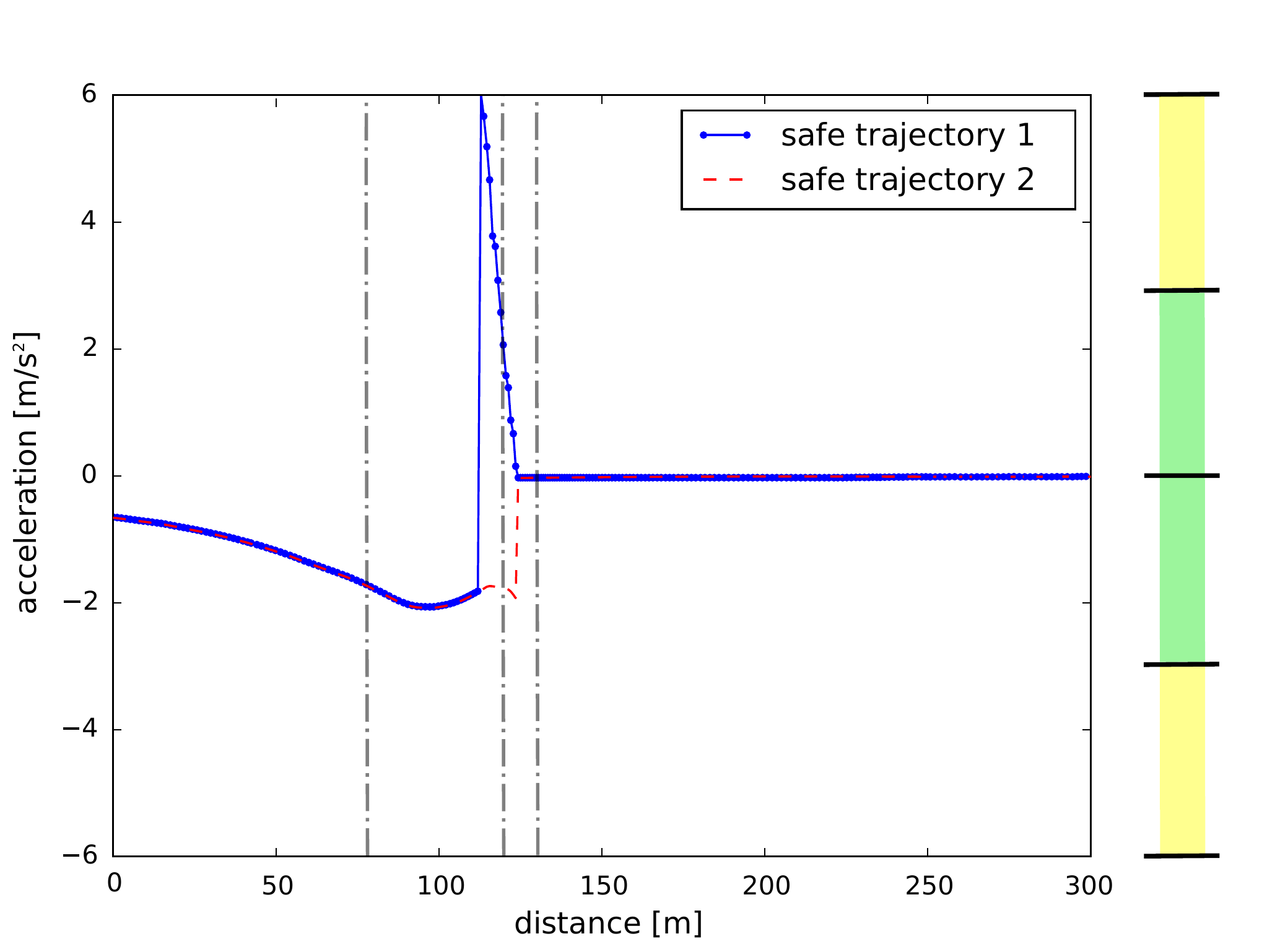}
      \llap{\includegraphics[trim={0cm 0cm 0cm 1.3cm},clip,width=0.97\linewidth]{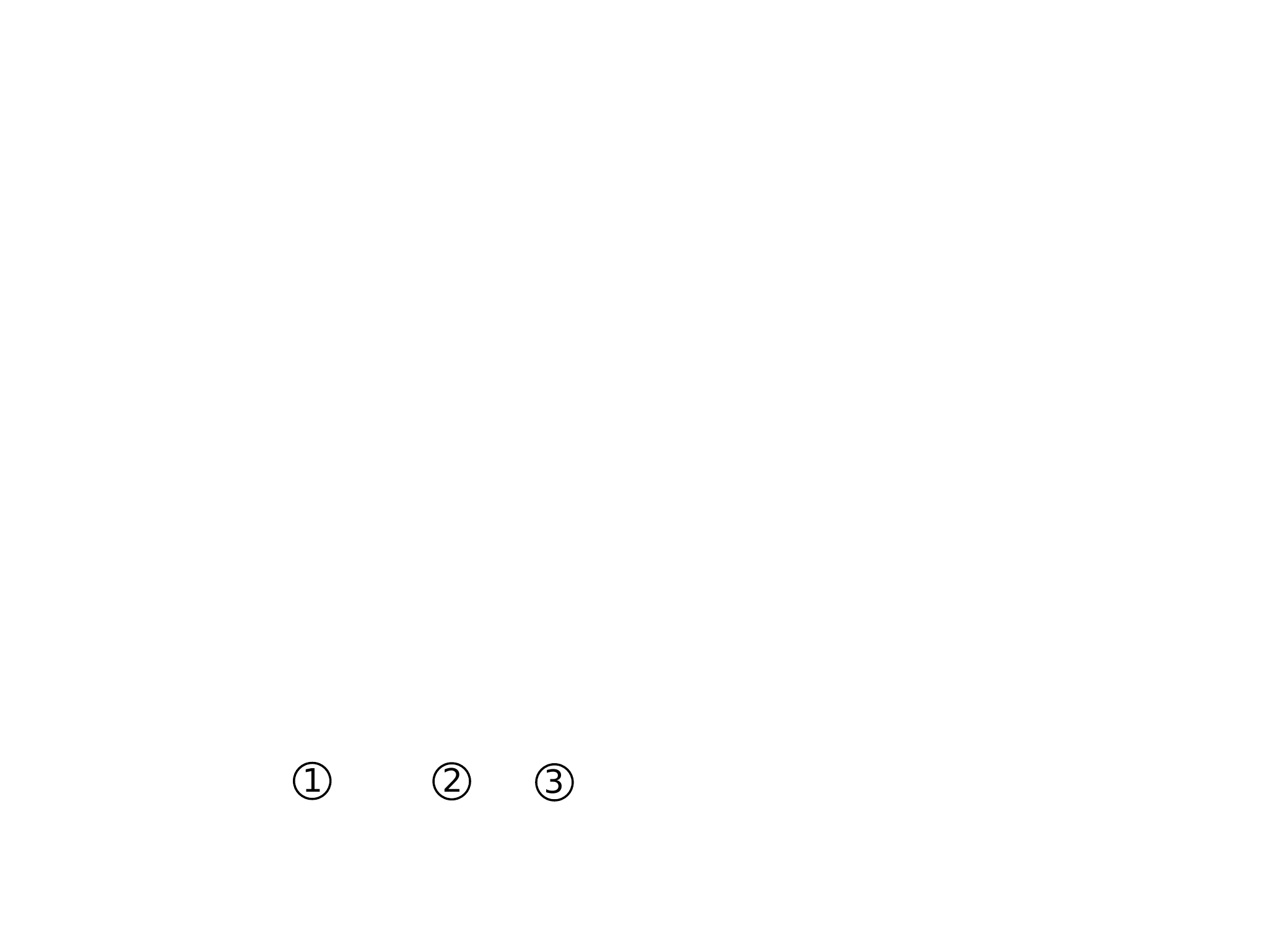}}      
      \caption{Trajectories for scenario with correct human behavior. Left: Acceleration plot (blue and red curve). Right: Comfortable and heavy intervention level.}
      \label{fig:acc1}
\end{figure}

\noindent point to pass in 
front of any potentially non-detectable vehicle $v_{\text{trg}}$ (black horizontal line).

Out of the concatenation of $a_{\text{const}}$, $a_{\text{stop}}$ and $a_{\text{acc}}$, two risk-minimizing/safe trajectories can be derived, which are plotted together
with the intervention levels in Fig. \ref{fig:acc1}. The three time steps are marked in the plot with three dashed vertical gray lines.

In this case the human behavior can be considered as being safe. The driver slows down until the intersection is mostly visible and
accelerates back to the desired cruising velocity. Consequently, it is always possible to safely stop at the stop line if a car emerges from the occluded area.

At $d=\unit[84] {m}$ the intersection is largely occluded, but the stop line is still $d_{\text{sl}} =  \unit[50] {m}$ away.
A deceleration of $a_{\text{stop}}= \unit[-1]{m/s^2}$ is proposed so that the velocity profile in the risk map avoids the risk spot.

When the ego vehicle is closer to the intersection with $d_{\text{sl}} =  \unit[15] {m}$ at $d=\unit[118] {m}$, the required deceleration increases to $a_{stop} = \unit[-2]{m/s^2}$.
At the same time, the intersection is visible to such an extent that an acceleration of $a_{\text{acc}}= \unit[6]{m/s^2}$ would allow to pass the intersection in front of the 
virtual car by reaching $v_{\text{trg}}=\unit[12]{m/s}$. Since $a_{\text{acc}}= \unit[6]{m/s^2}$ is in the heavy intervention level, $a_{\text{stop}}= \unit[-2]{m/s^2}$ is 
the recommended behavior.
If a real vehicle appears from the non-visible area, by executing $a_{\text{stop}}= \unit[-2]{m/s^2}$, the ego vehicle would be able to safely stop at the stop line without a collision.

Finally at $d=\unit[126] {m}$, the intersection is nearly completely visible. Keeping the velocity constant with $a_{\text{const}}=\unit[0]{m/s^2}$ would not result in collision anymore.
At all time steps the proposed behavior resembles the actual safe human behavior. The slope of one of the yellow curves matches the purple curve. 
Hence, the required deceleration $a_{\text{stop}}$ stays small and at no time a warning is triggered.

\subsection{Scenario with Critical Human Behavior}

\begin{figure}[H]
      \centering
      \framebox{\parbox{0.95\linewidth}{
      \centering
      \includegraphics[trim={0 5.54cm 0 0.1cm},clip,width=\linewidth]{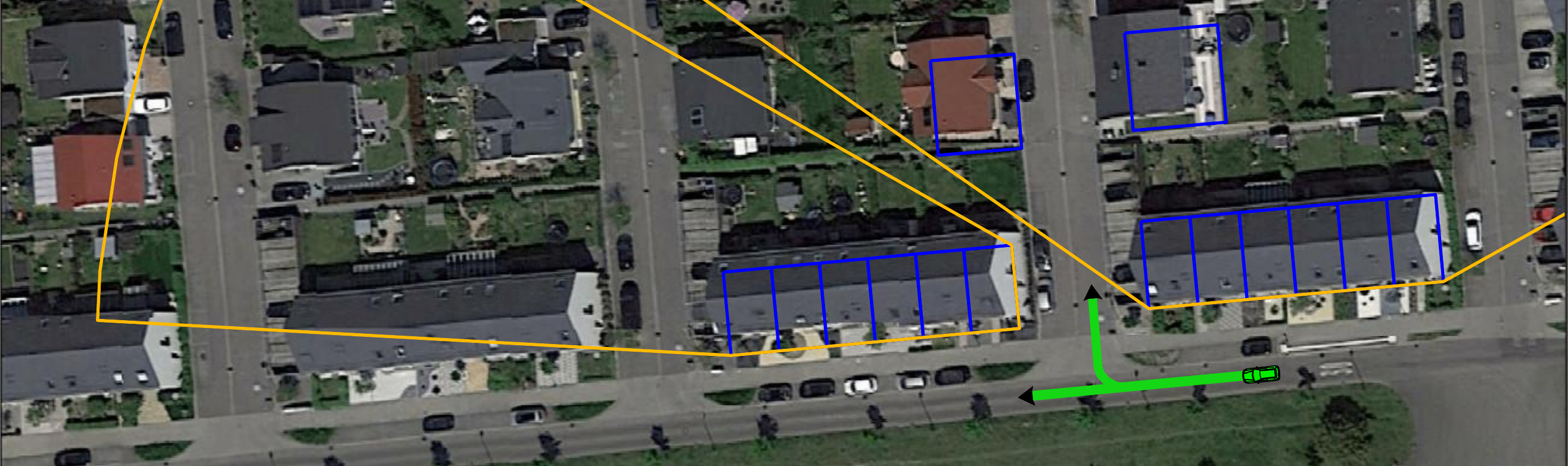}
      \includegraphics[trim={0 5.54cm 0 0.1cm},clip, width=\linewidth]{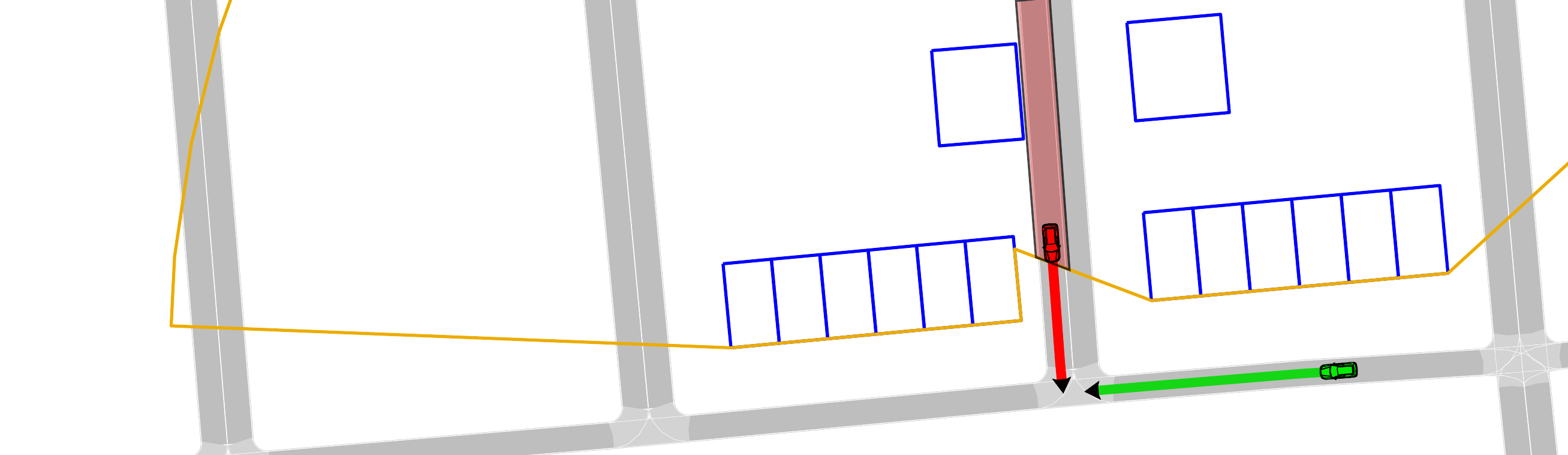}
      \includegraphics[trim={0.35cm 0cm 3cm 0cm},clip, width=0.97\linewidth]{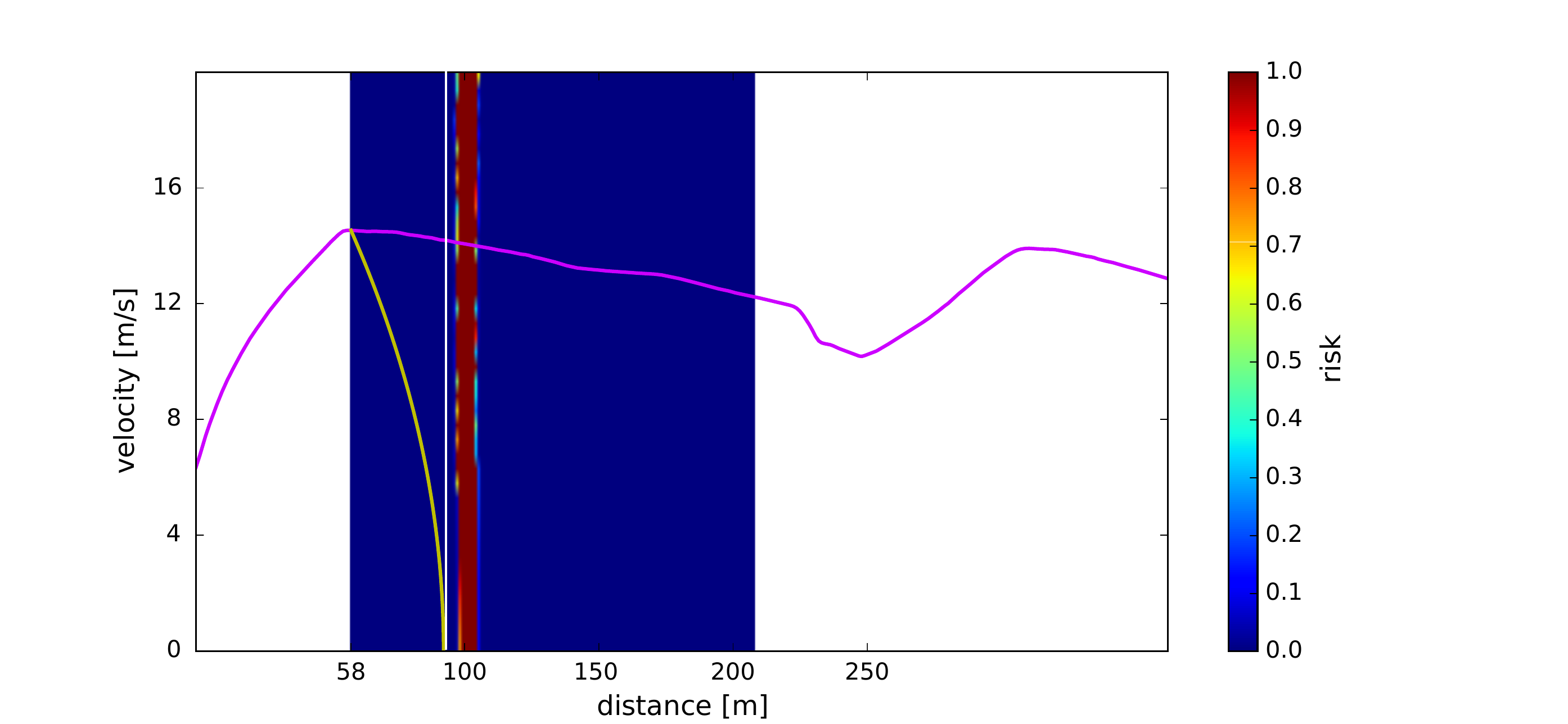}
      \llap{\includegraphics[trim={0.35cm 0cm 3cm 0cm},clip,width=0.97\linewidth]{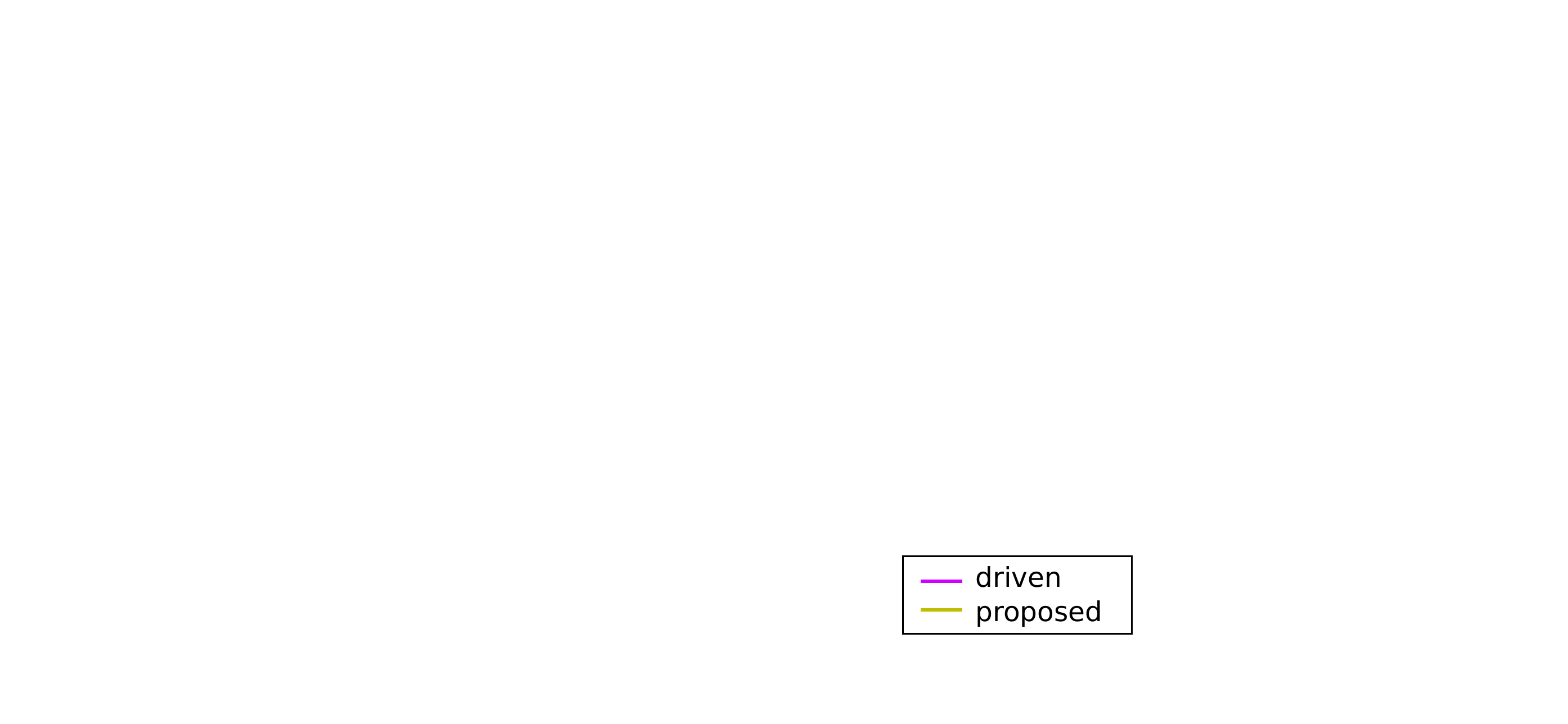}}
      \llap{\includegraphics[trim={0.35cm 0cm 3cm 0cm},clip,width=0.97\linewidth]{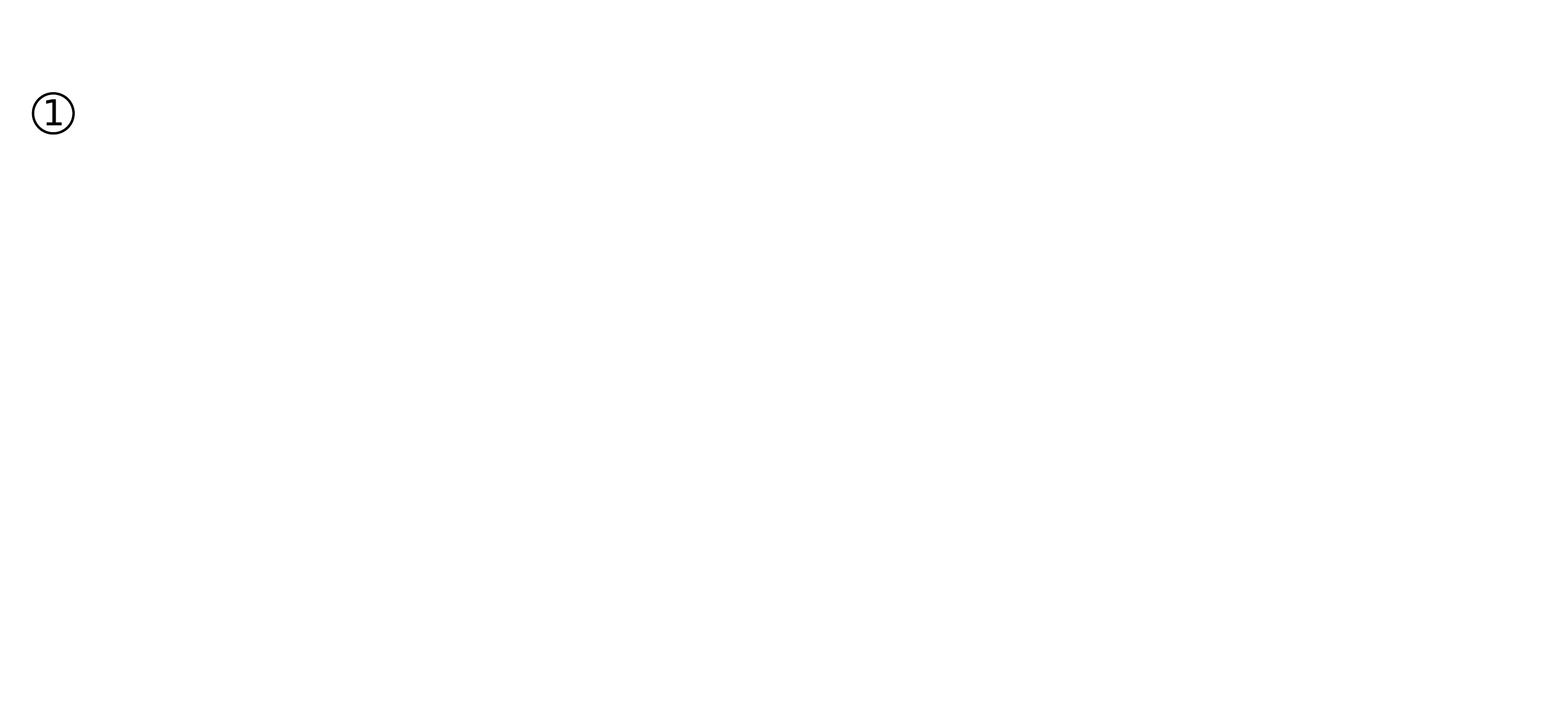}}   
      \includegraphics[trim={0 5.54cm 0 0.1cm},clip,width=\linewidth]{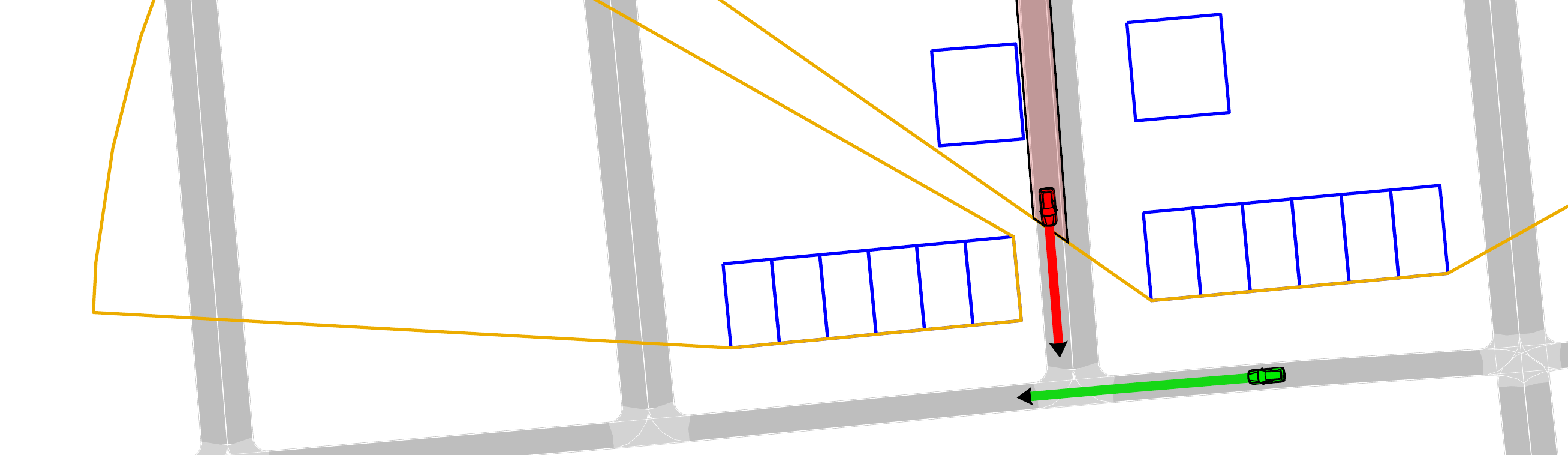}
      \includegraphics[trim={0.7cm 0cm 3cm 0cm},clip,width=0.97\linewidth]{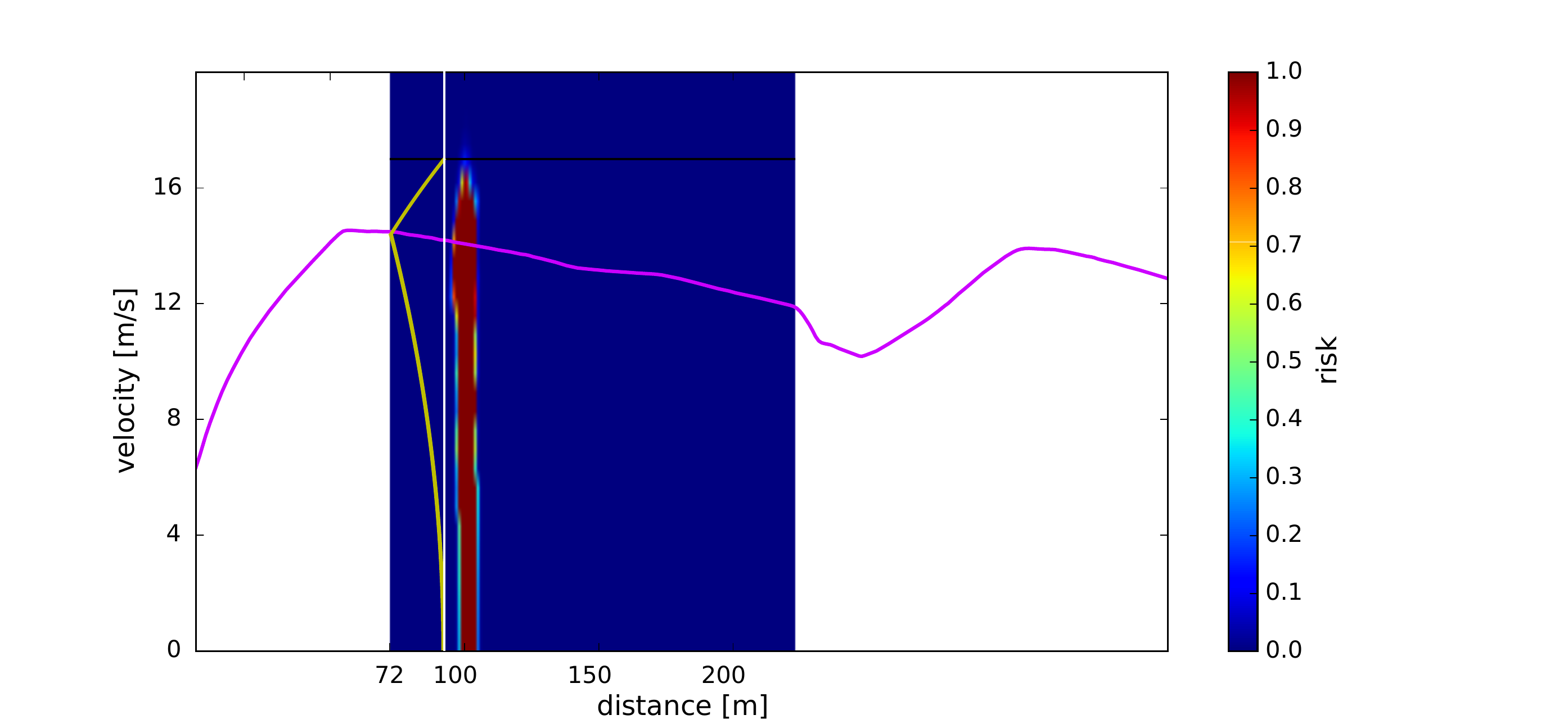}
      \llap{\includegraphics[trim={0.7cm 0cm 3cm 0cm},clip,width=0.97\linewidth]{images/fig12_4.pdf}}
      \llap{\includegraphics[trim={0.35cm 0cm 3cm 0cm},clip,width=0.97\linewidth]{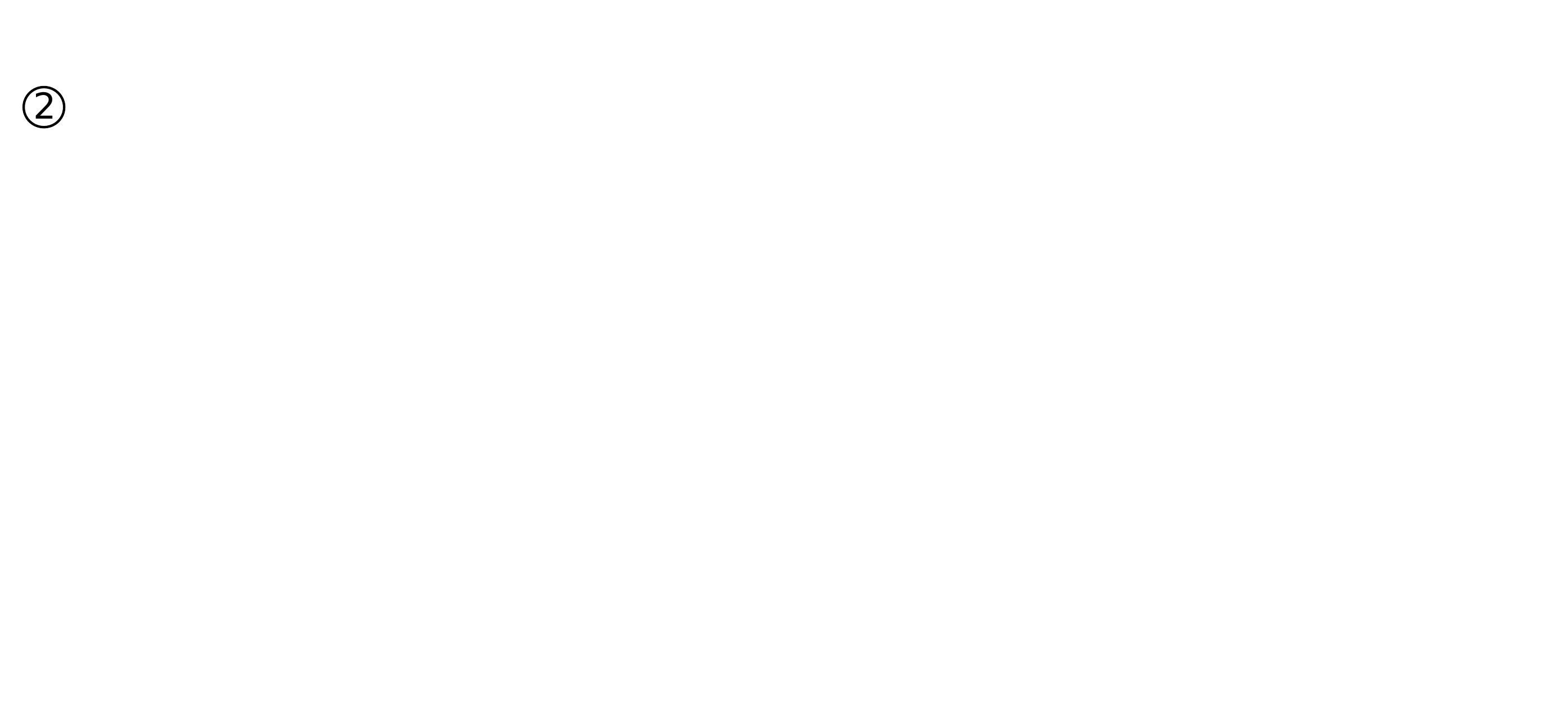}}   
      \includegraphics[trim={0 5.54cm 0 0.1cm},clip,width=\linewidth]{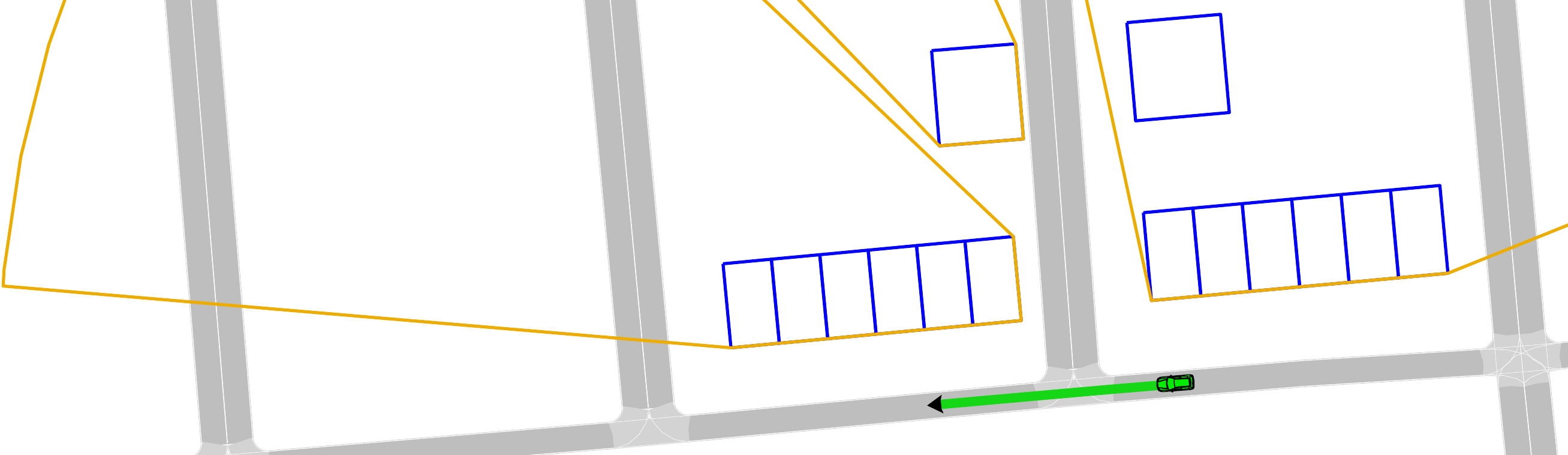}
      \includegraphics[trim={0.7cm 0cm 3cm 0cm},clip,width=0.97\linewidth]{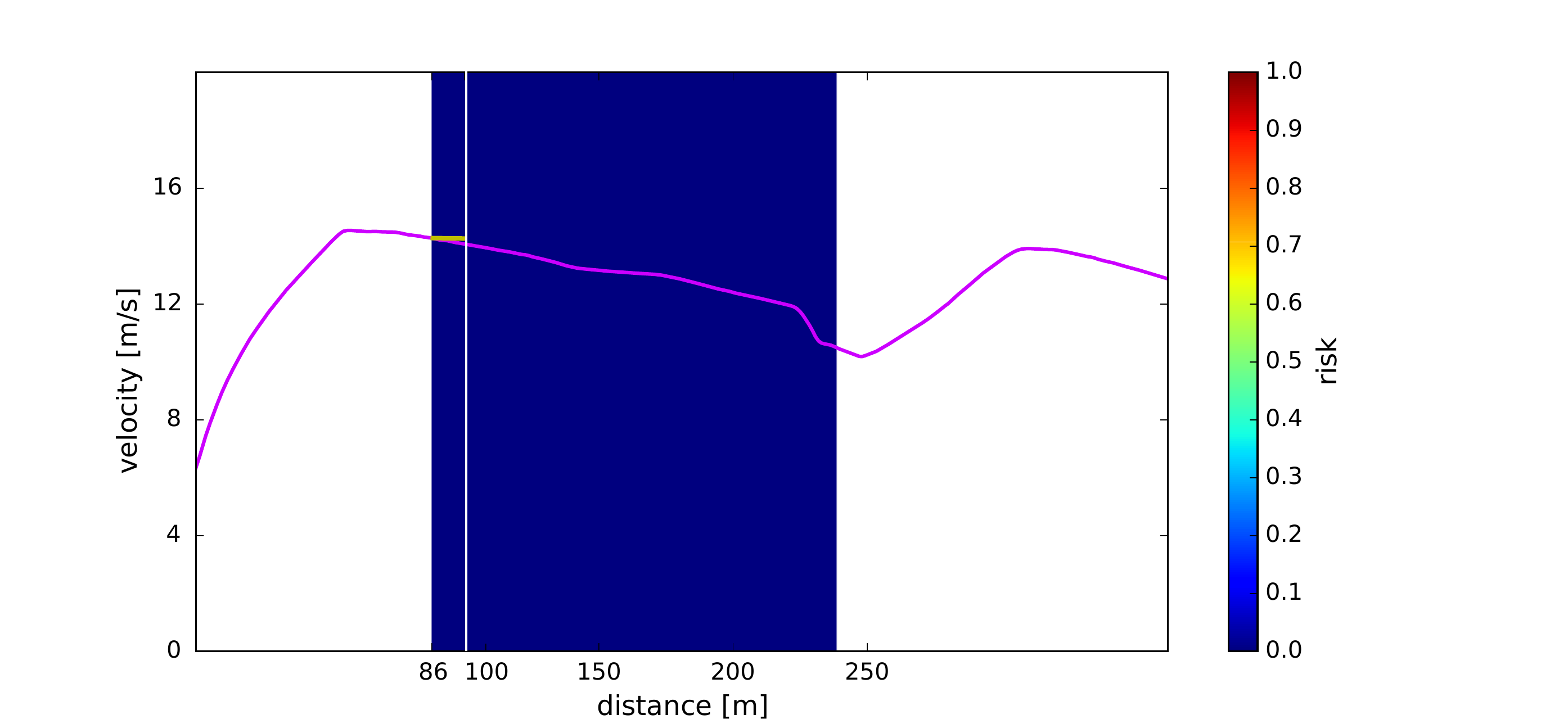}
      \llap{\includegraphics[trim={0.7cm 0cm 3cm 0cm},clip,width=0.97\linewidth]{images/fig12_4.pdf}}
      \llap{\includegraphics[trim={0.35cm 0cm 3cm 0cm},clip,width=0.97\linewidth]{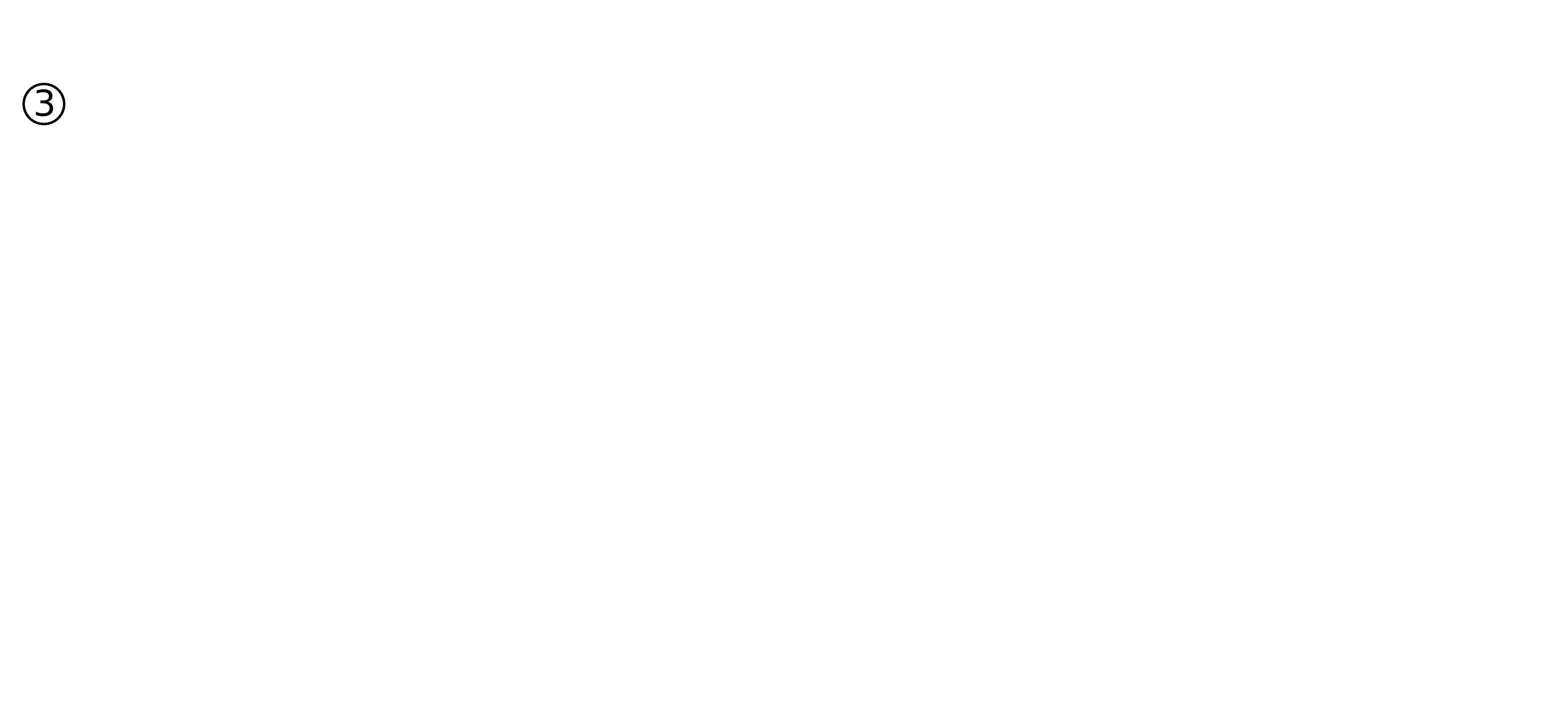}}   
      }}
      \caption{Evaluation of intersection assistance for limited visibility applied to a scenario with incorrect human behavior.}
      \label{fig:results_incorrect}
\end{figure}

\begin{figure}[H]
      \centering
      \includegraphics[trim={0 0cm 0 1cm},clip, width=\linewidth]{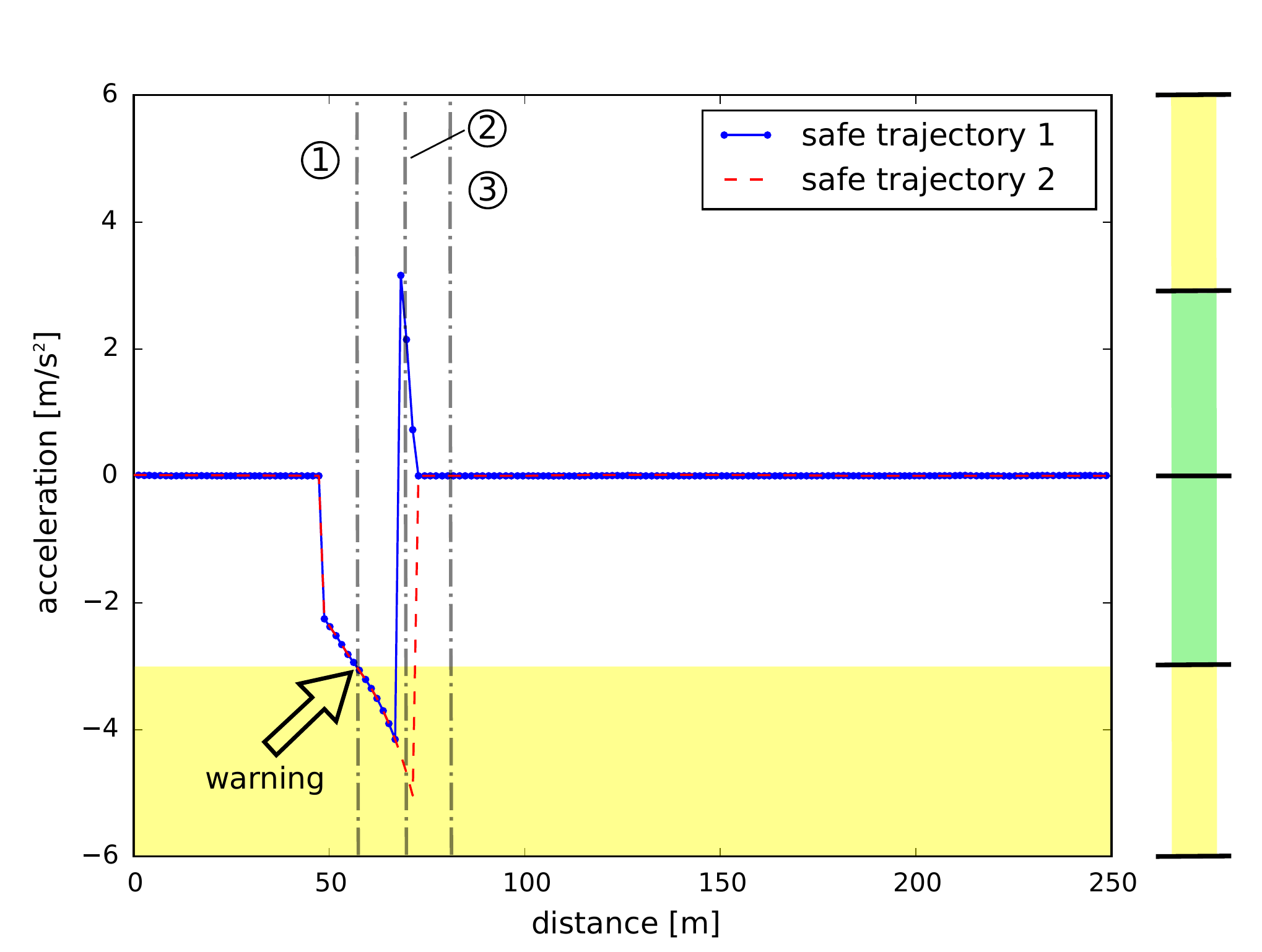}
      \caption{Trajectories for scenario with incorrect human behavior.} 
      \label{fig:acc2}
\end{figure}

Figs. \ref{fig:results_incorrect} and \ref{fig:acc2} depict the simulation results of 
the ego vehicle approaching an intersection with an occluded incoming lane with priority over the ego vehicle's lane for three time frames $d=[58,72,86] \unit[]{m}$. The driver does not reduce the speed

\noindent while coming closer to the intersection, which allows the assumptions that he is not considering potential hazards. Therefore, his behavior is seen as being critical.

At $d=58 \unit[]{m}$ the ego velocity has a maximal value of $v_{0} =  \unit[14] {m/s}$ so that a braking maneuver with $a_{\text{stop}}= \unit[-3]{m/s^2}$ is 
necessary to guarantee a safe halt in $d_{\text{sl}} =  \unit[30] {m}$. Because $a_{\text{stop}}=\unit[-3]{m/s^2}$ is the only behavior alternative and lies in the heavy intervention level, a warning is indicated to the driver. 

Shortly after, at $d=72 \unit[]{m}$ the braking distance decreases to $d_{\text{sl}} =  \unit[16] {m}$ whereas the velocity stays the same with $v_{0} =  \unit[14] {m/s}$, leading to $a_{\text{stop}}= \unit[-5]{m/s^2}$. At this point, it is easier to avoid the risk spot by accelerating to $v_{\text{trg}}=\unit[17]{m/s}$ with 
$a_{\text{acc}}= \unit[3]{m/s^2}$. Since $v_{\text{trg}}=\unit[17]{m/s}>\unit[50]{km/h}$, the acceleration $a_{\text{acc}}= \unit[3]{m/s^2}$ is neglected and the warning stays activated. In this context, $\unit[50]{km/h}$ is the allowed speed in inner-cities.

In this example, there is no real car on the right lane. 
The intersection becomes entirely visible at $d=86 \unit[]{m}$ and thus the risk map does not show a risk spot anymore.
At this point in time, driving with constant velocity $a_{\text{const}}=\unit[0]{m/s^2}$ is proposed to the driver without a warning.

In the first two time steps the behavior suggestion deviates from the actual human behavior. The slope of the yellow and purple curves have
different values. When $a_{\text{stop}}$ reaches the heavy intervention level, the system determines a critical driver behavior and gives a warning
about $3 \unit[]{s}$ before a possible crash.

\section{Discussion and Outlook} \label{sec:conclusion}
In this work, we introduced a novel intersection warning system that not only allows for the evaluation of collision risks from vehicles detected by on-board sensor,
but also of risks originating from hypothetical cars appearing at occlusions. 
The sensor's visibility area to look into the upcoming intersection for dynamic road entities is thus virtually enhanced and a risk-aversive behavior becomes plannable.

Simulations showed that the proposed system's behavior is matching the general behavior of a correctly acting human driver.
In scenarios in which the actual human behavior is differing from the proposed behaviors of the system, a warning is released
to successfully avoid potential hazards lying at areas which cannot be accessed by on-board sensor. 
Optionally, we display the safe trajectory with the lowest intervention level.

So that the system can be implemented for real-time use on a test car, the ego position has to be aligned on the R-LDM. This is achievable either with high-accurate GPS or a localization module. Furthermore, the provided map data in the R-LDM should contain detailed geometric information about the lanes and buildings. Only if collision risk of other real vehicles is also considered, active sensors (e.g. lidar or camera) are needed to retrieve the relative position of those vehicles.

Currently, occlusions only result from buildings around the intersection. The sensor range can additionally be reduced by 
other static objects, such as parked cars and trees on the side, or by cars driving nearby. 
Furthermore, the assumed behavior of the virtual entity matches the behaviors of cars, but not that of pedestrians or bicyclists.
Nevertheless, it is straightforward to extend the proposed general approach for other occlusion sources and entity types.

When planning necessary braking and accelerating maneuvers, human factors, like driving experience, have not been integrated yet.
At the same time, one or more proposed behavior alternatives are possibly not executable since they would violate traffic rules. 
Future research will concentrate on incorporating realistic driver types, so that a personalization of warning can be achieved.

\section*{Acknowledgment}
\noindent This work has been supported by the European Union’s Horizon 2020 projects VI-DAS (grant agreement number 690772) and
INLANE (grant agreement number 687458).
The authors would like to thank Daniela Aguirre Salazar, Daan De Geus, and Cristhiam Felipe Pulido Riveros for their
support.

\ifCLASSOPTIONcaptionsoff
  \newpage
\fi

\bibliographystyle{IEEEtran}
\bibliography{bib}




\end{document}